\documentclass[10pt,twocolumn,letterpaper]{article}

\usepackage{cvpr}
\usepackage{times}
\usepackage{epsfig}
\usepackage{graphicx}
\usepackage{amsmath}
\usepackage{amssymb}

\usepackage{siunitx}
\usepackage{bm} % optional
\usepackage{array}
\usepackage{multirow}
\usepackage{graphicx}
\usepackage{float}
\usepackage{pgfplots}
\usepackage{pgfplotstable}
\pgfplotsset{compat=1.11,
        /pgfplots/ybar legend/.style={
        /pgfplots/legend image code/.code={%
        \draw[##1,/tikz/.cd,bar width=7pt,yshift=-0.4em,bar shift=0pt]
                plot coordinates {(0cm,0.8em)};},
},
}
\usepgfplotslibrary{colorbrewer}
\pgfplotsset{width=10cm,compat=1.9,cycle list/Paired-10}
\usepackage[T1]{fontenc}
\usepackage[utf8]{inputenc}
\usepackage{authblk}
\usepackage{enumitem}

\usepackage[breaklinks=true,bookmarks=false]{hyperref}

\newtheorem{theorem}{Theorem}

\cvprfinalcopy % *** Uncomment this line for the final submission

\def\cvprPaperID{1156} % *** Enter the CVPR Paper ID here

\ifcvprfinal\fi

\begin{document}

%%%%%%%%% TITLE
\title{Event Probability Mask (EPM) and Event Denoising Convolutional Neural Network (EDnCNN) for Neuromorphic Cameras}

\author[1]{R. Wes Baldwin}
\author[2]{Mohammed Almatrafi}
\author[1]{Vijayan Asari}
\author[1]{Keigo Hirakawa}
\affil[1]{Department of Electrical Engineering, University of Dayton}
\affil[2]{Department of Electrical Engineering, Umm Al-Qura University

{\tt\small \{baldwinr2, vasari1, khirakawa1\}@udayton.edu, mmmatrafi@uqu.edu.sa}}

\renewcommand\Authands{ and }

\maketitle
\thispagestyle{empty}

%%%%%%%%% ABSTRACT
\begin{abstract}
%Neuromorphic camera sensors overcome the speed and dynamic range limitations in conventional cameras without the high cost, size, and computational requirements of high-speed cameras. Unfortunately, one must tolerate high levels of noise to operate them at maximum sensitivity. Existing event denoising methods rely on hand-crafted features due to a lack of labeled event data. 
This paper presents a novel method for labeling real-world neuromorphic camera sensor data by calculating the likelihood of generating an event at each pixel within a short time window, which we refer to as ``event probability mask'' or EPM. Its applications include (i) objective benchmarking of event denoising performance, (ii) training convolutional neural networks for noise removal called ``event denoising convolutional neural network'' (EDnCNN), and (iii) estimating internal neuromorphic camera parameters. We provide the first dataset (DVSNOISE20) of real-world labeled neuromorphic camera events for noise removal.
\end{abstract}

%%%%%%%%% BODY TEXT
\section{Introduction} \label{intro}

Neuromorphic (a.k.a.~event-based) cameras offer a hardware solution to overcome limitations of conventional cameras, with high temporal resolution (${>}$\SI{800}{\kilo\hertz}), low latency (\SI{20}{\micro\second}), wide dynamic range (\SI{120}{\dB}), and low power ($10{-}$\SI{30}{\milli\watt})~\cite{lichtsteiner2008128}. This is accomplished by a dynamic vision sensor (DVS), which reports the log-intensity changes (i.e. events) of each pixel in microseconds. However, performance of methods using neuromorphic cameras deteriorate with noise. This fact has been cited as a major challenge in recent research~\cite{czech2016evaluating,gallego2019event}. Noise is noticeable in low light conditions, where events triggered by minor intensity fluctuations dominate over the usable signal in the scene. Currently, there is no reliable way to benchmark the denoising performance because the exact distribution of noise in DVS circuitry---which is environment-, scene-, and sensor-dependent---is still unknown. Since neuromorphic cameras generate millions of events each second, it is impractical to manually label each event. This has precluded machine learning approaches to event denoising until this point.

\begin{figure}[t]
    \centering
    \includegraphics[width=1.0\linewidth]{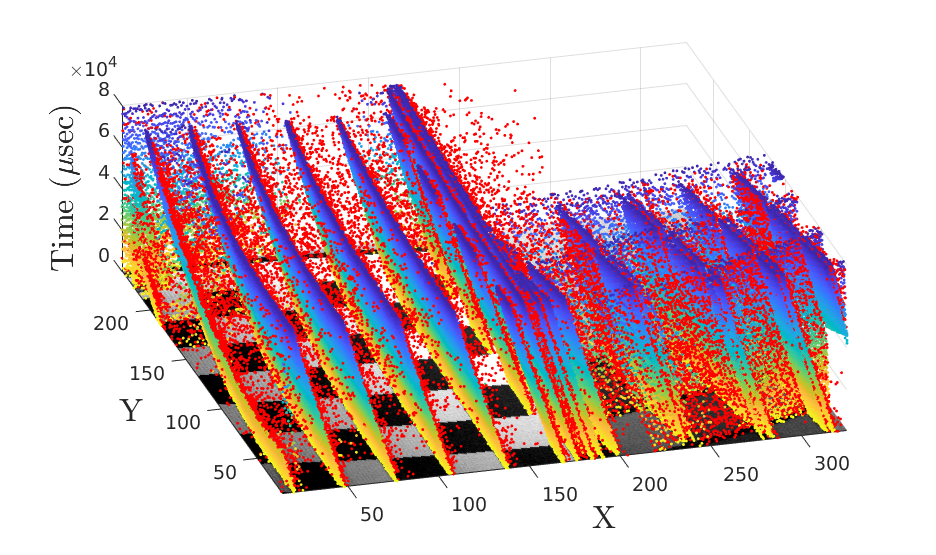}
   \caption{Proposed EDnCNN denoising applied to ``CheckerFast'' sequence in DVSNOISE20 dataset. DVS dots are colored by time and overlaid on APS image. Red dots were classified by EDnCNN as noise. EDnCNN characterizes real-world noise distribution by learning the mapping between actual and noise-free DVS events.}
\label{fig:surfacePlot}
\end{figure}

We propose the notion of an ``event probability mask'' (EPM) -- a label for event data acquired by real-world neuromorphic camera hardware. We infer the log--likelihood probability of an event within a neuromorphic camera pixel by combining the intensity measurements from active pixel sensors (APS) and the camera motion captured by an inertial measurement unit (IMU). Our contributions are:% as follows:
\begin{itemize}[leftmargin=5.5mm]
  \item {\bf Event Probability Mask (EPM):} spatial-time neuromorphic event probability label for real-world data;
  \item {\bf Relative Plausibility Measure of Denoising (RPMD):} objective metric for benchmarking DVS denoising;
  \item {\bf Event Denoising CNN (EDnCNN):} DVS feature extraction and binary classifier model for denoising; 
  \item {\bf Calibration:} a maximum likelihood estimation of threshold values internal to the DVS circuitry; and
  \item {\bf Dataset (DVSNOISE20):} labeled real-world neuromorphic camera events for benchmarking denoising.
\end{itemize}

\begin{figure}[t]
    \centering
    \includegraphics[width=1.0\linewidth]{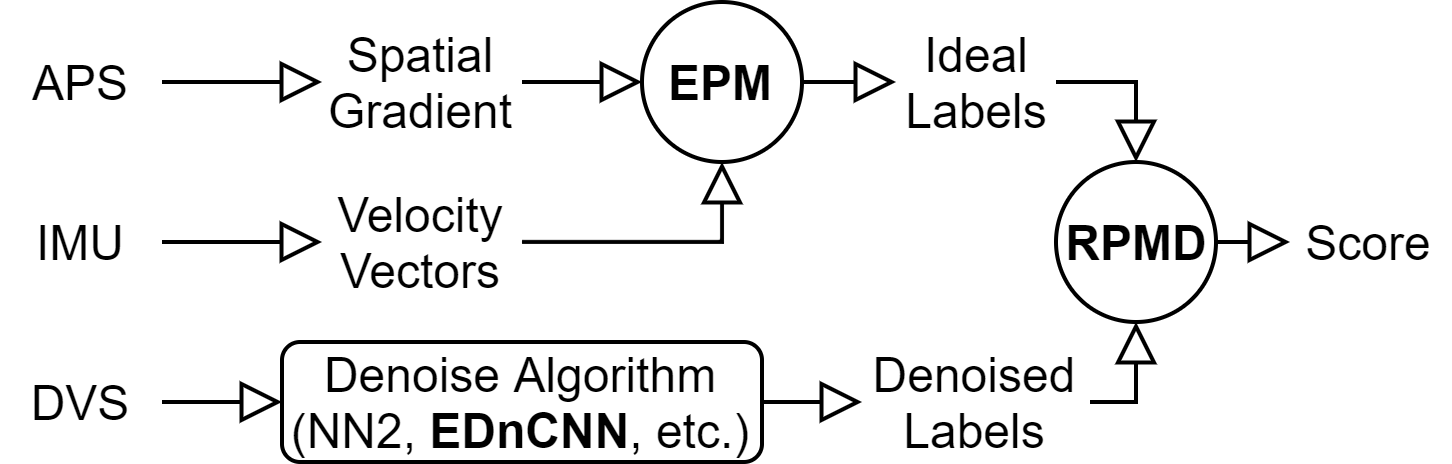}
   \caption{EPM is a prediction of idealized DVS behavior -- computed from spatial gradients (APS) and velocities (IMU). RPMD benchmarks performance by comparing denoising labels to EPM, which acts as a proxy for unobservable noise-free DVS events.}
\label{fig:overviewChart}
\end{figure}

%This paper is organized as follows. Section~\ref{sec:bkgnd} reviews background and related work on denoising. Section~\ref{eventProbability} details EPM and labeling of DVS data using APS and IMU. Applications of EPM are described in Section~\ref{sec:benchmark} (RPS for benchmarking), Section~\ref{eventDenoising} (EDnCNN for denoising), and Section~\ref{sec:estCameraParams} (threshold calibration). New benchmarking dataset and benchmarking results are shown in Section~\ref{experiments}. Concluding remarks are made in Section~\ref{conclusion}.

%outlines a benchmarking technique for quantitative assessment of denoising algorithms. Section~\ref{eventDenoising} develops the proposed event denoising neural network for maximizing the log likelihood probability of the denoised events. Section~\ref{estCameraParams} describes the proposed calibration technique for estimating the thresholding values internal to the DVS circuitry. 

\section{Background and Related Work}
\label{sec:bkgnd}

\subsection{Neuromorphic Cameras}

An APS imaging sensor synchronously measures intensity values observed by the photodiodes to generate frames. Although APS is a mature technology generating high quality video, computer vision tasks such as object detection, classification, and scene segmentation are challenging when the sensor or the target moves at high-speed (i.e. blurring) or in high dynamic range scenes (i.e. saturation). High-speed cameras rely on massive storage and computational hardware, making them unsuitable for real-time applications or edge computing.

A DVS is an asynchronous readout circuit designed to determine the precise timing of when log-intensity changes in each pixel exceed a predetermined threshold. Due to their asynchronous nature, DVS events lack the notion of frames. Instead, each generated event reports the row/column pixel index, the timestamp, and the polarity. Log-intensity change is a quantity representing relative intensity contrast, yielding a dynamic range far wider than a conventional APS. Typical event cameras have a minimum threshold setting of $15$--$50\%$ illumination change---with the lower limit determined by noise~\cite{gallego2019event}.

\begin{figure}[t]
\centering
\includegraphics[width=1.0\linewidth]{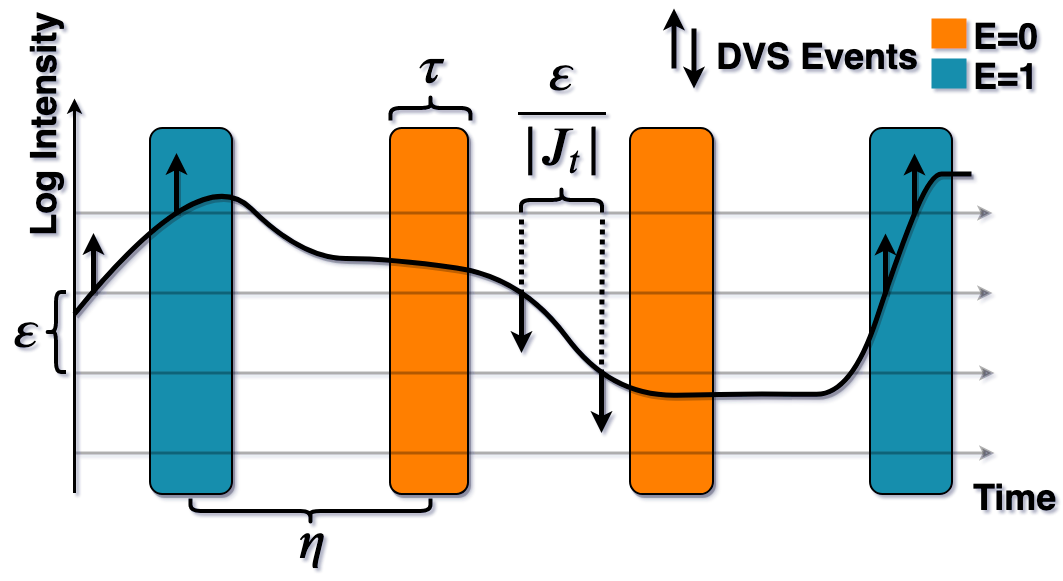}
   \caption{DVS events are generated when the log-intensity $J$ exceeds a predefined threshold $\varepsilon$. APS frames are exposed for $\tau$ seconds, occurring at the rate of $\eta$ seconds. Based on scene content and camera motion, EPM label predicts whether an event would have occurred ($E=1$) or not ($E=0$) during APS exposures.}
\label{fig:eventGen}
\end{figure}

In this work, we make use of a dynamic active vision sensor (DAVIS) that combines the functionality of DVS and APS~\cite{brandli2014240}. Two read-out circuits share the same photodiode, operating independently. One outputs the log-intensity changes; the other records the linear intensity at up to 40+ frames per second. In addition, the DAVIS camera has an IMU, operating at the 1kHz range with timestamps synchronized to the APS and DVS sensors. See Figure~\ref{fig:eventGen}.

Since their introduction, neuromorphic cameras have proven useful in simultaneous localization and mapping (SLAM)~\cite{rebecq2016evo,weikersdorfer2013simultaneous}, optical flow~\cite{almatrafi2019davis,benosman2012asynchronous,zhu2018ev}, depth estimation~\cite{chaney2019learning,zhu2019unsupervised}, space applications~\cite{cheung2018probabilistic,cohen2018approaches}, tactile sensing~\cite{naeini2019novel,rigi2018novel}, autonomous navigation~\cite{maqueda2018event,sanket2019evdodge}, and object classification~\cite{baldwin2019inceptive,barua2016direct,bi2019graph,gehrig2019end,orchard2015hfirst}. Many approaches rely on hand-crafted features such as~\cite{chin2019event,lagorce2016hots,lenz2018high,mueggler2017fast,sironi2018hats}, while other applications use deep learning architectures trained using simulated data~\cite{rebecq2019events,scheerlinck2019ced}. 

\subsection{Denoising} \label{relatedWork}

There are largely four types of random noise in neuromorphic cameras. First, an event is generated even when there is no real intensity change. Referred to as ``background activity'' (BA), these false alarms severely impact algorithm accuracy and consume bandwidth. Second, an event is not generated, despite an intensity change (i.e. ``holes'' or false negatives). Third, the timing of the event arrival is stochastic. Lastly, although proportional to the edge magnitude (e.g. high contrast change generates more events than low contrast), the actual number of events for a given magnitude varies randomly.

Most existing event denoising methods are concerned with removing BA---examples include bioinspired filtering~\cite{barrios2018less} and hardware-based filtering~\cite{khodamoradi2018n,liu2015design}. Spatial filtering techniques leverage the spatial redundancy of pixel intensity changes as events tend to correspond to the edges of moving objects. Thus, events are removed due to spatial isolation~\cite{czech2016evaluating,delbruck2008frame} or through spatial-temporal local plane fitting~\cite{benosman2013event}. Similarly, temporal filters exploit the fact that a single object edge gives rise to multiple events proportional in number to the edge magnitude. Temporal filters remove events that are temporally redundant~\cite{alzugaray2018asynchronous} or ambiguous. Edge arrival typically generates multiple events. The first event is called an ``inceptive event'' (IE)~\cite{baldwin2019inceptive} and coincides with the edge's exact moment of arrival. Events directly following IE are called ``trailing events'' (TE), representing edge magnitude. TE have greater ambiguity in timing because they occur some time after the edge arrival.

\begin{figure*}[t]
\centering
\begin{tabular}{m{0.01\textwidth}m{0.165\textwidth}m{0.165\textwidth}m{0.165\textwidth}m{0.165\textwidth}m{0.165\textwidth}}
\rotatebox{90}{DAVIS Sensor} & \includegraphics[width=1.07\linewidth]{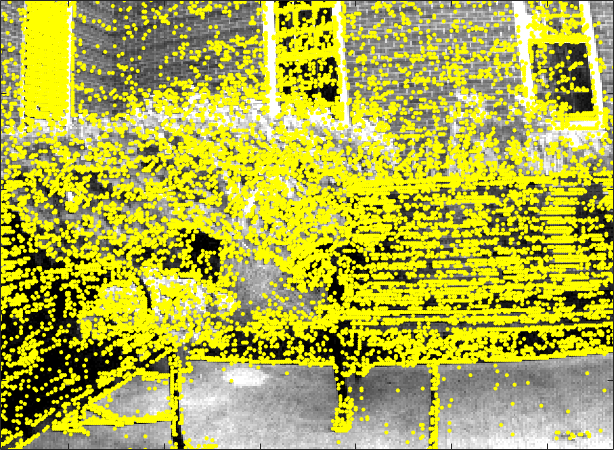} & \includegraphics[width=1.07\linewidth]{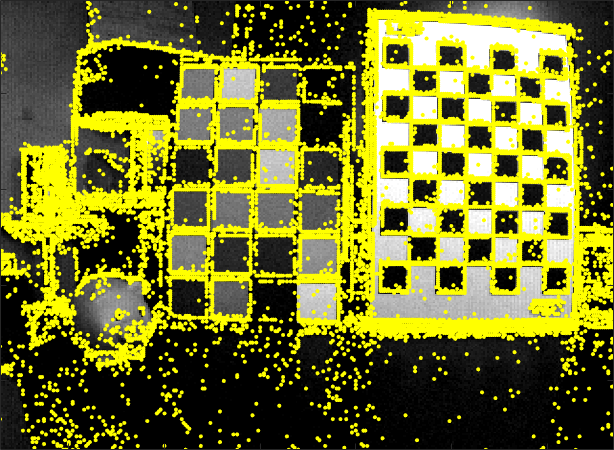} & \includegraphics[width=1.07\linewidth]{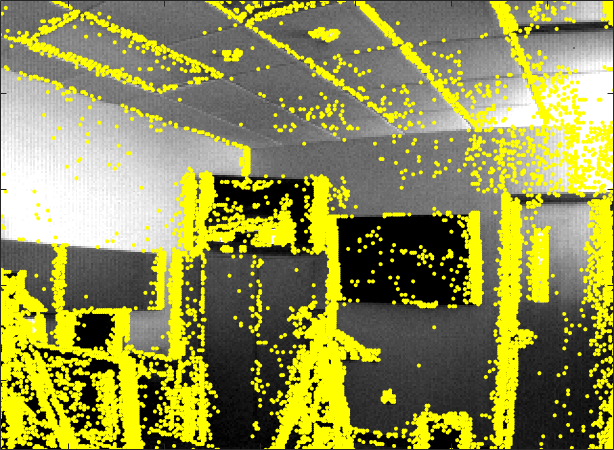} & 
\includegraphics[width=1.07\linewidth]{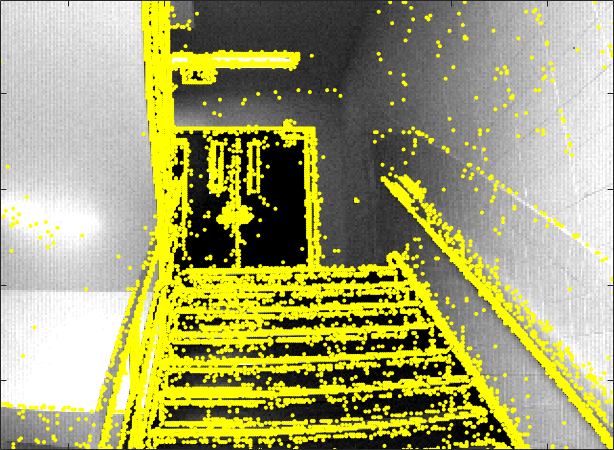} & \includegraphics[width=1.07\linewidth]{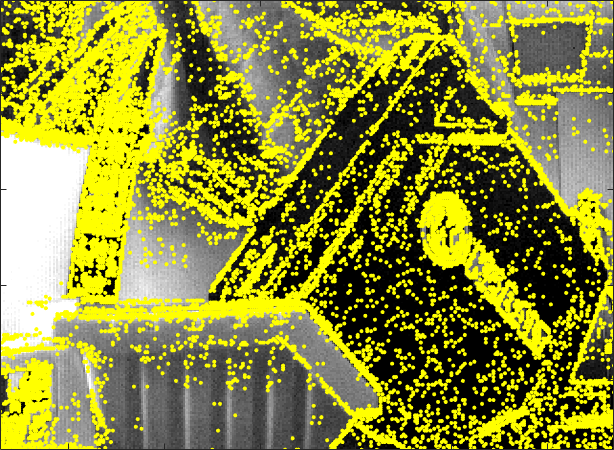} \\
% \rotatebox{90}{Image} & \includegraphics[width=1.07\linewidth]{pics/scenes3/3_472_image.png} & \includegraphics[width=1.07\linewidth]{pics/scenes3/8_299_image.png} & \includegraphics[width=1.07\linewidth]{pics/scenes3/16_255_image.png} & 
% \includegraphics[width=1.07\linewidth]{pics/scenes3/13_296_image.png} & \includegraphics[width=1.07\linewidth]{pics/scenes3/14_259_image.png} \\
\rotatebox{90}{\textbf{EPM}} & \includegraphics[width=1.07\linewidth]{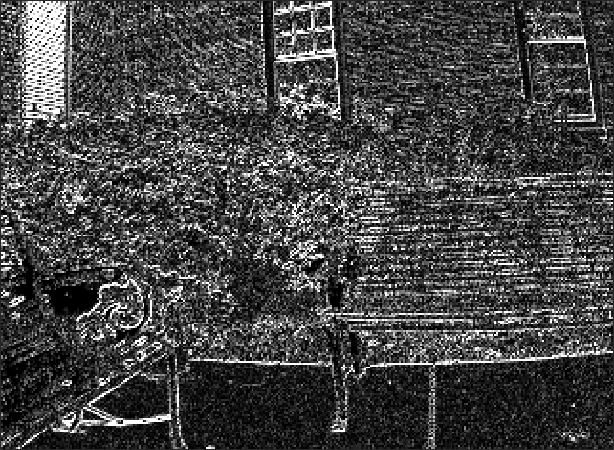} & \includegraphics[width=1.07\linewidth]{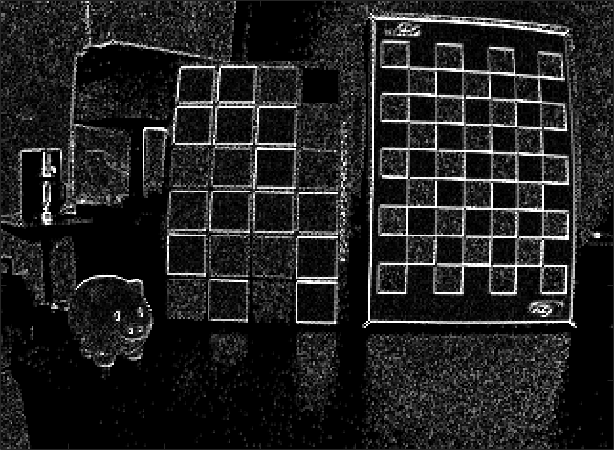} & 
\includegraphics[width=1.07\linewidth]{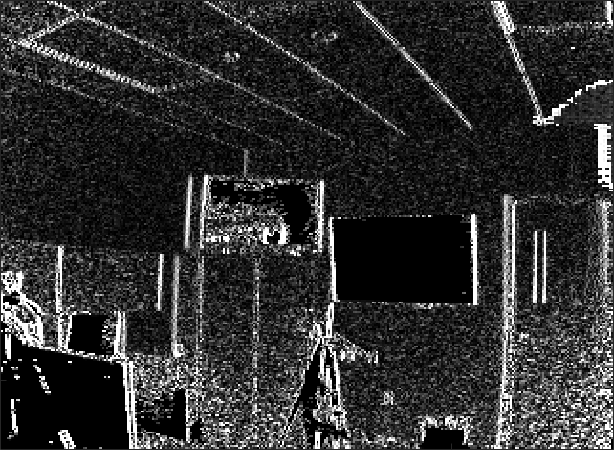} & 
\includegraphics[width=1.07\linewidth]{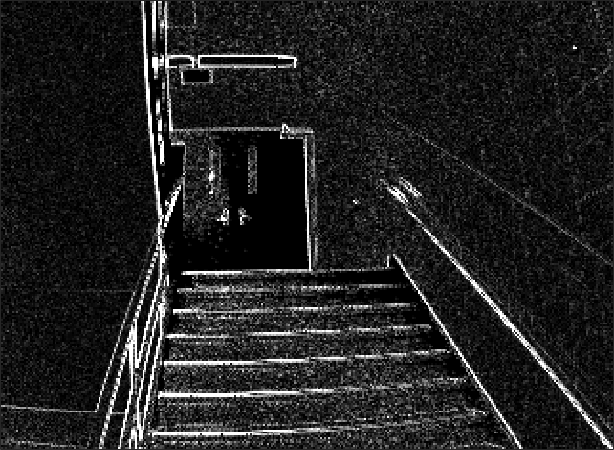} & \includegraphics[width=1.07\linewidth]{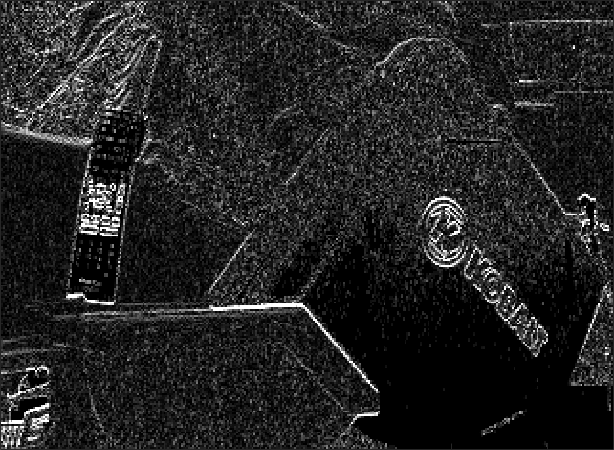} \\
\multicolumn{1}{r}{} & \multicolumn{1}{c}{Benches} & \multicolumn{1}{c}{CheckerSlow} & \multicolumn{1}{c}{LabFast} & \multicolumn{1}{c}{Stairs} & \multicolumn{1}{c}{Toys}
\end{tabular}%
\caption{(First Row) Examples from DVSNOISE20 dataset. Noisy (raw) DVS events overlayed on APS frames. (Second Row) Proposed event probability mask (EPM) predicting the noise-free DVS behavior. Intensity values denote probability 0 (black) -- 1 (white).}
\label{tab:dataset_EPM}
\end{figure*}

\subsection{Neuromorphic Camera Simulation}
\label{sec:simulation}

Simulators such as ESIM~\cite{rebecq2018esim} and PIX2NVS~\cite{bi2017pix2nvs} artificially generate plausible neuromorphic events corresponding to a user-specified input APS image or 3D scene. The simulated neuromorphic events have successfully been used in machine learning methods to perform tasks such as motion estimation~\cite{bryner2019event,sekikawa2019eventnet,stoffregen2019event} and event-to-video conversion~\cite{rebecq2019events}. However, the exact probabilistic distribution of noise within neuromorphic cameras is the subject of ongoing research, making accurate simulation challenging. To the best of our knowledge, we are unaware of any prior denoising methods leveraging the exact and explicit characterization of the DVS noise distribution.

\section{Event Probability Mask} \label{eventProbability}

%We first develop the requisite mathematics for DAVIS sensor hardware (Section~\ref{sec:davis}), optical flow (Section~\ref{sec:OF}, and inertial measurement unit (Section~\ref{sec:IMU}). Then EPM is derived in Section~\ref{sec:label}.

%A denoising method yielding filtered events that maximize this log likelihood probability is one that is most consistent with the intensity changes of the ground truth scenery.

We describe below a novel methodology of predicting the behavior of the DVS from APS intensity measurements and IMU camera motion. We derive the likelihood probability of an event in DVS pixels of a noise-free camera, a notion we refer to as ``event probability mask'' (EPM). EPM serves as a proxy for ground truth labels. For example, EPM identifies which of the real-world events generated by actual DVS hardware are corrupted by noise, thereby overcoming the challenges associated with modeling or simulating the noise behavior of DVS explicitly (see Section~\ref{sec:simulation}).

%One application to EPM is benchmarking of the performances of event-based denoising algorithms based on real-world DVS data (see Section~\ref{sec:benchmark}). We also propose an objective quality metric for DVS data called relative plausibility score (RPS) derived from EPM. Neural networks can be trained to remove noisy event data by learning to maximizing the RPS (see Section~\ref{sec:denoising}). Neural network-based denoising methods can be trained on labeled real-world DVS data, implicitly learning the noise distribution of the actual neuromorphic camera hardware rather than relying on simulated noise. EPM can also be used for calibration of parameters internal to the neuromorphic cameras (see Section~\ref{sec:calibration}).

\subsection{Proposed Labeling Framework}
\label{sec:label}

Let $I:\mathbb{Z}^2 \times \mathbb{R}\to \mathbb{R}$ denote an APS video (the signal), where $I(\bm{X},t)$ is the radiance at pixel $\bm{X}\in\mathbb{Z}^2$ and at time $t\in\mathbb{R}$. The log amplifier in DVS circuit yields a log-intensity video $J:\mathbb{Z}^2\times \mathbb{R}\to \mathbb{R}$, modeled as: 
\begin{align}\label{eq:J}
J(\bm{X},t):=\log(a I(\bm{X},t)+b),
\end{align}
where $a$ and $b$ are the gain and offset, respectively. In \emph{noise-free neuromorphic camera hardware}, idealized events are reported by DVS when the log-intensity exceeds a predefined threshold $\varepsilon>0$:
\begin{align}\label{eq:events}
t_i(\bm{X})&:=\arg\min_{t}\notag\\
&\left\{t>t_{i-1}(\bm{X})\Big| |J(\bm{X},t)-J(\bm{X},t_{i-1}(\bm{X}))| \geq \varepsilon \right\}\notag\\
p_i(\bm{X})&:=\operatorname{sign}(J(\bm{X},t_i(\bm{X}))-J(\bm{X},t_{i-1}(\bm{X}))).
\end{align} 

% \begin{equation}
% \begin{aligned}\label{eq:events}
% d_i(\bm{X},t):={}&J(\bm{X},t)-J(\bm{X},t_{i-1}(\bm{X}))\\
% t_i(\bm{X}):={}&\arg\min_{t}\left\{t>t_{i-1}(\bm{X})\Bigg| |d_i(\bm{X},t)| \geq \varepsilon \right\}\\
% p_i(\bm{X}):={}&\operatorname{sign}(d_i(\bm{X},t_i)).
% \end{aligned} 
% \end{equation}
                
Ideally, each reported event from noise-free neuromorphic camera hardware provides the spatial location $\bm{X}$, precise timestamp $t_i$ that $J(\bm{X},t)$ crosses the threshold, and polarity $p_{i}\in\{+1,-1\}$ indicating whether the change in the log-pixel intensity was brighter or darker respectively (see Figure~\ref{fig:eventGen}).

Now suppose $\{t_i(\bm{X}),p_i(\bm{X})\}$ refers to a set of events obtained from \emph{real, practical, noisy DVS hardware}. We consider formalizing the DVS event denoising as a hypothesis test of the form:
\begin{align}\label{eq:hypothesis1}
    \begin{cases}
        H_0:  &|J(\bm{X},t_{i}(\bm{X}))-J(\bm{X},t_{i-1}(\bm{X}))| \geq \varepsilon \\
        H_1:  &|J(\bm{X},t_{i}(\bm{X}))-J(\bm{X},t_{i-1}(\bm{X}))| < \varepsilon. 
    \end{cases}
\end{align}
That is, we would like to determine whether an event as described by $(t_i,p_i)$ corresponds to an actual temporal change in the log-pixel intensity exceeding the threshold $\varepsilon$. However, this formalism has a major disadvantage; the hypothesis test on $(t_i,p_i)$ relies on another event, $(t_{i-1},p_{i-1})$ which may also be noisy. Thus in this work, we revise the hypothesis test as follows:
\begin{align}\label{eq:hypothesis2}
    &\begin{cases}
        H_0:  &\text{$t_i(\bm{X})\in[t,t+\tau)$ for some $i$}\\
        H_1:  &\text{$t_i(\bm{X})\notin[t,t+\tau)$ for all $i$},
    \end{cases}
\end{align}
where $\tau$ is a user-specified time interval (set to the integration window of APS in our work; see Theorem~\ref{thm:Mx} below). Notice that the new hypothesis test decouples $(t_i,p_i)$ from $(t_{i-1},p_{i-1})$. Hypothesis test in \eqref{eq:hypothesis2} also abstracts away the magnitude and timing noises, while faithfully modeling BA and holes.

Define the event probability mask (EPM) $M:\mathbb{Z}^2\times\mathbb{R}\to[0,1]$ as the Bernoulli probability of null hypothesis:
\begin{align}\label{eq:EPM}
\begin{split}
    M(\bm{X}, t):=&Pr[H_0]= Pr[\exists i \text{ s.t. } t_i(\bm{X})\in[t,t+\tau)].
\end{split}
\end{align}
Intuitively, {\bf EPM quantifies the plausibility of observing an event within the time window $[t,t+\tau)$}. If an event occurred during $[t,t+\tau)$ but $M(\bm{X},t_i)\approx 0$, then this is an implausible event, likely caused by noise. On the other hand, if $M(\bm{X},t_i)\approx 1$, we have a high confidence that the event corresponds to an actual temporal change in the log-pixel intensity exceeding $\varepsilon$. In this sense, EPM is a proxy for soft ground truth label---the inconsistencies between EPM and the actual DVS hardware output identify the events that are corrupted by noise. EPM $M:\mathbb{Z}^2\times\mathbb{R}\to[0,1]$ can be computed from APS and IMU measurements as explained in Theorem~\ref{thm:Mx} below. 
\begin{theorem}\label{thm:Mx}
Let  $\bm{\theta}(t)=(\theta_x(t),\theta_y(t),\theta_z(t))^T$ represent the instantaneous 3-axis angular velocity of camera measured by IMU's gyroscope. Let $A:\mathbb{Z}^2\times\mathbb{Z}\to\mathbb{R}$ denote APS measurements with exposure time $\tau$. Assume camera configuration with focal length $f$, principal point  $c_x,c_y$, and skew parameter $\kappa$. Then
\begin{align}\label{eq:Mx_probability}
    M(\bm{X},t)
    =\begin{cases}
    \frac{\tau|J_t(\bm{X},t)|}{\varepsilon}&\text{if $|J_t(\bm{X},t)|<\frac{\varepsilon}{\tau}$}\\
    1&\text{else}.
    \end{cases}
\end{align}
where  $J_t(\bm{X},t)$ is as described in \eqref{eq:J_t} and \eqref{eq:V}.
\end{theorem} 
Proof and derivation is provided in Appendix~\ref{sec:proof}. Examples of EPM are shown in Figure~\ref{tab:dataset_EPM}. Note that this method requires two parameters internal to the DAVIS camera, namely the threshold value $\varepsilon>0$ and the offset value $O\in\mathbb{R}$. Calibration procedure to obtain these values is described in Section~\ref{sec:estCameraParams}.

\subsection{Limitations}
\label{sec:limitations}

%The proposed event probability mask is a novel framework for labeling event data from real-world neuromorphic camera hardware. However, we acknowledge that there are some practical limitations to EPM.

%The APS framerate is very slow compared to DVS, making it difficult to predict DVS behavior unless the underlying motion of the scene is slow---ideally subpixel motion per frame. In Theorem~\ref{thm:Mx}, we overcame this limitation by leveraging the IMU hardware, which operates in the kilohertz range. This allows us to handle pixel motions that are orders of magnitude faster. Specifically, as explained in the proof of Theorem~\ref{thm:Mx} in Appendix~\ref{sec:proof}, we infer the pixel velocity stemming from camera motion using the gyroscopes inside IMU. 

%Computing EPM is possible for static scenes (i.e.~only camera movement) and if the camera motion is rotational. 

\begin{figure}[t]
\centering
    \includegraphics[width=1.0\linewidth]{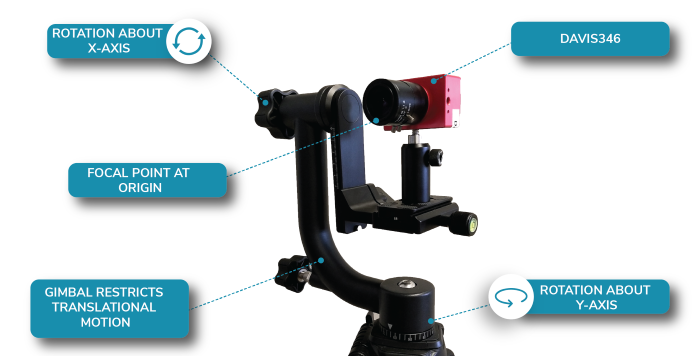}
   \caption{Camera setup for DVSNOISE20 collection. Gimbal limits camera motion while centering the focal point at the origin.}
\label{fig:gimbal}
\end{figure}

EPM calculation requires static scenes (i.e.~no moving objects) and rotation-only camera motion (i.e.~no translational camera movement) to avoid occlusion errors. We address this issue by acquiring data using a camera configuration as shown in Figure~\ref{fig:gimbal} (additional details in Section~\ref{sec:data}). We emphasize, however, this is only a limitation for benchmarking and network training (Section~\ref{sec:training}). These restrictions are removed at inference because spatially global and local pixel motions behave similarly within the small spatial window used by our denoising model. Our empirical results in Section~\ref{sec:otherexp} confirm robustness to such model assumptions. Additionally, Theorem~\ref{thm:Mx} is only valid for constant illumination (e.g.~fluorescent light flicker is detectable by DVS). Moreover, since the dynamic range of the DVS is much larger than the APS, it is not possible to calculate an EPM for pixels at APS extremes. Examples shown in Figure~\ref{tab:dataset_EPM}. 

\begin{figure*}
\begin{align}\label{eq:J_t}
    J_t(\bm{X})\approx& -\frac{\tau \nabla A(\bm{X},t)}{ A(\bm{X},t)-O}
    \begin{pmatrix}
    |V_x(\bm{X},t)|&0\\0&|V_y(\bm{X},t)|
    \end{pmatrix}
\bm{V}(\bm{X},t)\\
\label{eq:V}
     \bm{V}(\bm{X},t)
    =&\begin{pmatrix}f&\kappa&c_x\\0&f&c_y\end{pmatrix}
    \begin{pmatrix}
    0&-\theta_z(t)&\theta_y(t)\\
    \theta_z(t)&0&-\theta_x(t)\\
    -\theta_y(t)&\theta_x(t)&0
    \end{pmatrix}
    \begin{pmatrix}
    f&\kappa&c_x\\0&f&c_y\\0&0&1
    \end{pmatrix}^{-1}
    \begin{pmatrix}x\\y\\1\end{pmatrix}
\end{align}
%\hline
\noindent\rule{\textwidth}{0.5pt}
\end{figure*}

Another limitation is that the value of $M(\bm{X})$ diminishes when the camera motion is \emph{very} slow. This is due to the fact that the events are infrequently generated by DVS, reducing the probability of observing an event within a given time window $[t,t+\tau)$. While EPM captures this phenomenon accurately (limited only by the IMU sensitivity), it is difficult to discriminate noisy events from events generated by extremely slow motion.

\section{Application: Denoising Benchmarking}
\label{sec:benchmark}

EPM is the first-of-its-kind benchmarking tool to enable quantitative evaluation of denoising algorithms against real-world neuromorphic camera data. Given a set of events $(t_i,p_i)$, let $E:\mathbb{Z}^2\to\mathbb\{0,1\}$ denote an event indicator:
\begin{align}
    E(\bm{X},t)=\begin{cases}
    1 & \text{if $t_i(\bm{X})\in[t,t+\tau)$ for some $i$}\\
    0 & \text{if $t_i(\bm{X})\notin[t,t+\tau)$ for all $i$}.
    \end{cases}
\end{align}
Then if EPM is known, the log-probability of the events $(t_i,p_i)$ is explicitly computable:
\begin{align}\label{eq:logPE}
    &\log Pr[E] = \\
    &\sum_{\bm{X}\in\mathbb{Z}^2} E(\bm{X})\log M(\bm{X})+(1-E(\bm{X}))\log (1-M(\bm{X})).\notag
\end{align}
(Proof: for each pixel $\bm{X}\in\mathbb{Z}^2$, $\log Pr[E(\bm{X})=1]=M(\bm{X})$ and $\log Pr[E(\bm{X})=0]=1-M(\bm{X})$.) This log-probability can be used to assess the level of noise present in real, practical, noisy DVS hardware. On the other hand, if $(t_i',p_i')$ denotes the outcome of an event denoising method, then the corresponding log-probability $\log Pr[E']$ is an objective measure of the denoising performance. The improvement from noisy events $(t_i,p_i)$ to denoised events $(t_i',p_i')$ can be quantified by $\log Pr[E']-\log Pr[E]$.

The objective of denoising methods, then, is to yield a set of events $(t_i',p_i')$ to maximize $\log Pr[E']$. In fact, the theoretical bound for best achievable denoising performance is computable. It is
\begin{multline}
 \max_{E:\mathbb{Z}^2\to\mathbb\{0,1\}} \log Pr[E]=\\
 \sum_{\bm{X}\in\mathbb{Z}^2}\log \max(M(\bm{X}),1-M(\bm{X})),
\end{multline}
which is achieved by a thresholding on $M(\bm{X})$
\begin{align}\label{eq:E_opt}
E_{opt}(\bm{X})=
\begin{cases}
1&\text{if $M(\bm{X})>0.5$}\\
0&\text{if $M(\bm{X})\leq 0.5$}.
\end{cases}
\end{align}
(Proof: if $M(\bm{X})\leq 1-M(\bm{X})$ then $M(\bm{X})\leq 0.5$.) Thus we propose an objective DVS quality metric for denoising called ``relative plausibility measure of denoising'' (RPMD), defined as
\begin{equation}
RPMD:=\frac{1}{N}\log\frac{Pr[E_{opt}]}{Pr[E]}
\end{equation}
%\begin{align}
%\begin{split}
%&RPS:=\frac{1}{N}\log\frac{Pr[E_{opt}]}{Pr[E]}=\log %Pr[E_{opt}]-\log Pr[E]\\
%&=\frac{1}{N}\sum_{\bm{X}\in\mathbb{Z}^2}\Big( \log \max( %M(\bm{X}),1-M(\bm{X}))\\
%&~~-E(\bm{X})\log M(\bm{X})-(1-E(\bm{X}))\log %(1-M(\bm{X}))\Big),
%\end{split}
%\end{align}
where $N$ is the total number of pixels. Lower RPMD values indicate better denoising performance with 0 representing the best achievable performance. Benchmarking results using RPMD are shown in Figures~\ref{fig:benchmark} and \ref{fig:simNoise}.

\begin{figure}[t]
\centering
    \includegraphics[width=1.0\linewidth]{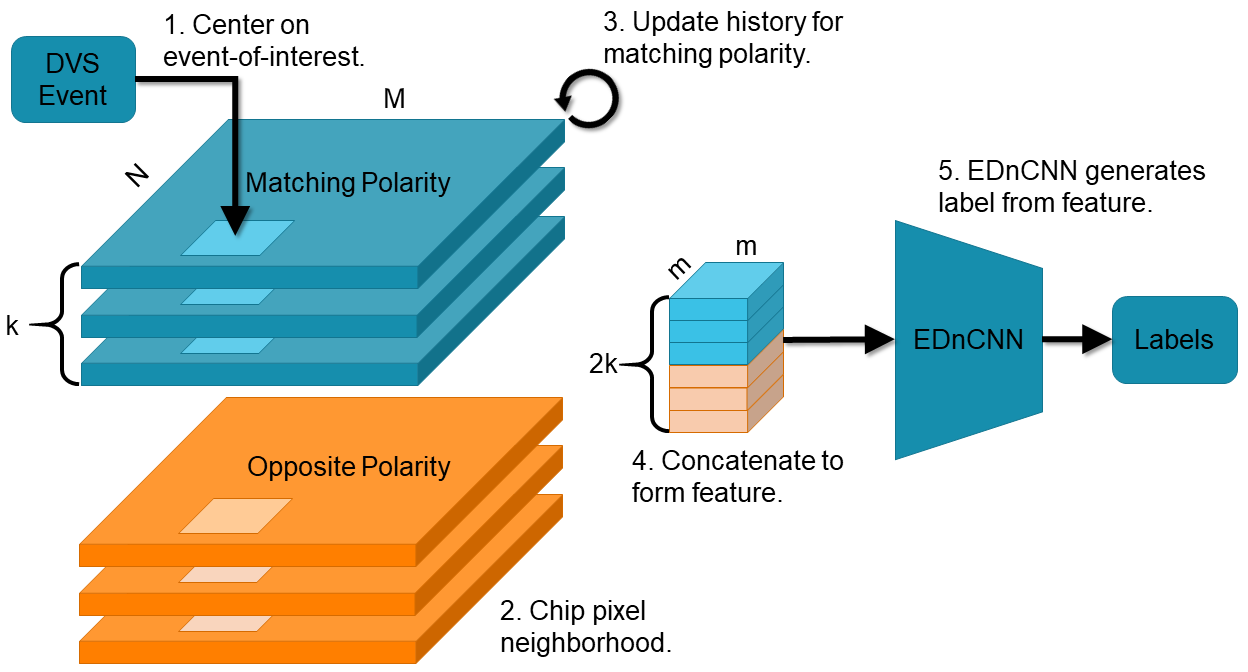}
   \caption{Multiple time-surfaces are generated from $k$ most recent events in $m\times m$ neighborhood of event-of-interest $\bm{X}$. All surfaces are concatenated and passed to EDnCNN. EDnCNN performs binary classification to yield a denoising label.}
\label{fig:featureGen}
\end{figure}

%\begin{figure*}[t]
%\centering
%\includegraphics[width=1.0\linewidth]{pics/featureGen3DSurf.png}
%   \caption{When an event at $\bm{X}$ (event-of-interest) is received, the most recent $k$ events from its spatial local neighbors are used to generate multiple time-surfaces ($k=2$ shown here). This is done for both polarities and all surfaces are concatenated before being passed to EDnCNN. EDnCNN then executes binary classification to make an inference on whether the event-of-interest is noisy or real.}
%\label{fig:featureGen}
%\end{figure*}

\section{Application: Event Denoising CNN} \label{eventDenoising}

Event denoising for neuromorphic cameras is a binary classification task whose goal is to determine whether a given event corresponds to a real log-intensity change or noise. We propose EDnCNN, an event denoising method using convolutional neural networks. EDnCNN is designed to carry out the hypothesis test in \eqref{eq:hypothesis2}, where the null hypothesis $H_0$ states that the event $t_i(\bm{X})$ is expected to be generated by DVS within a short temporal window $[t,t+\tau)$ due to changes in log-intensity.

The input to EDnCNN is a 3D vector generated only from DVS events. The training data is composed of DVS and the corresponding EPM label. Once trained, EDnCNN does not require APS, IMU, stationary scenes, or restricted camera motion. By training on DVS data from actual hardware, EDnCNN benefits from learning noise statistics of real cameras in real environments.

\subsection{Input: Event Features} \label{sec:efeatures}

There exist a wide array of methods to extract features from neuromorphic camera data~\cite{alzugaray2018asynchronous,bi2019graph, chandrapala2016invariant, gehrig2019end, gehrig2019eklt, lagorce2015spatiotemporal, lagorce2016hots, orchard2015hfirst, sironi2018hats}. These methods
%, centered around solving global problems, 
are designed to summarize thousands or even millions of events into a single feature to carry out high-level tasks such as
%This approach is ideal for high-level tasks such as 
object tracking, detection, and classification. However, event denoising is a low-level classification task. Denoising requires inference on pixel-level signal features rather than high-level abstraction of scene content. For example, there is a high likelihood that an isolated event is caused by noise, whereas spatially and temporally clustered events likely correspond to real signal~\cite{padala2018noise}. For this reason, event denoising designed to discriminate IE, TE, and BA would benefit from features that faithfully represent the local temporal and spatial consistency of events.

In denoising, we take inspiration from PointConv~\cite{wu2019pointconv}, a method for generating nonlinear features using local coordinates of 3D point clouds. Unlike PointConv, designed for three continuous spatial domains, a DVS event is represented by two discrete spatial and one continuous temporal dimension. We leverage the discrete nature of the spatial dimensions by mapping the temporal information from the most recently generated event at each pixel to construct a time-surface~\cite{lagorce2016hots}. This is similar to FEAST~\cite{afshar2019event}, which extracts features from a spatial neighborhood of time-surface near the event-of-interest. However, the temporal history of recent events at each pixel is averaged into a single surface, obfuscating the spatial consistency of event timing useful to denoising.

%% start major edit

Combining ideas from PointConv and FEAST, we propose to encode the events within the spatial-temporal neighborhood of the event-of-interest $(t_i(\bm{X}),p_i(\bm{X}))$. The EDnCNN input is a feature vector  $Q\in\mathbb{R}^{m\times m\times k\times 2}$. Here, $m\times m$ refers to the size of the spatial neighborhood centered at pixel $\bm{X}$ where the event-of-interest occurred.  At each pixel within this spatial neighborhood, we wish to encode $k$ most recently occurring events (i.e.~before $t_i(\bm{X})$) of polarities $p_i=-1$ and $p_i=1$ (hence the dimension $2$ in $Q$). Note that the temporal neighborhood is not thresholded by time but by the number of events. This allows for automatic adaptation to pixel velocity---events corresponding to a slower moving edge require a longer temporal window than a fast moving edge with very frequent event reporting.

When an event $t_i(\bm{X}),p_i(\bm{X})$ is received, we populate $Q(:,:,1,-1)$ and $Q(:,:,1,+1)$ with $m\times m$ relative time-surfaces formed by the difference between the time stamp of event-of-interest $t_i(\bm{X})$ and the time stamps of the most recent events at each neighborhood pixel with the polarities $-1$ and $+1$ respectively. We repeat the construction of time-surfaces using the second most recent events at each neighborhood pixel, which is stored into $Q(:,:,2,-1)$ and $Q(:,:,2,+1)$ etc., until $k$ most recent events at every pixel are encoded into $Q$. See Figure~\ref{fig:featureGen}. As a side note, encoding of EDnCNN input feature is very memory efficient. Each time-surface is the size of the DVS sensor, meaning the overall memory requirement is $M\times N \times k\times 2$ where $(M,N)\in\mathbb{Z}^2$ is the spatial resolution of the DVS sensor, storing the most recent $k$ events. In our implementation, $m$ was set to 25, $k$ was 2, and the resolution of DAVIS346 is $(M,N)=346\times260$.

\subsection{Network Architecture}

Let $\widehat{E}_{\phi}:\mathbb{Z}^2\to\{0,1\}$ be the output of the EDnCNN binary classifier, where the network coefficients is denoted $\phi$. The output $\widehat{E}_{\phi}(\bm{X})=1$ implies that an event $(t_i,p_i)$ corresponds to a real event. Testing showed a shallow network could quickly be trained using the features described in Section~\ref{sec:efeatures}, which is ideal for high-performance inference. EDnCNN is composed of three $3\times3$ convolutional layers (employing ReLU, batch normalization, and dropout) and two fully connected layers. The learning was performed via Adam optimizer with a decay rate of 0.1 and a learning rate of 1E-4. The network is trained on EPM-labeled DVS events generated during APS frame exposure. Once trained, EDnCNN can classify events at any time. Due to the fact that small local patches are primarily scene independent, EDnCNN can perform well against new scenes and environments and \emph{requires no tuning or calibration} at inference.

\subsection{Training} 
\label{sec:training}
We present three strategies for training EDnCNN and prove that these strategies are statistically equivalent, given sufficient training data volume. The first approach aims at minimizing the RPMD by maximizing a related function:
\begin{align}\label{eq:cost1}
    \phi_{opt1}=&\arg \max_{\phi} \sum_{\bm{X}} Pr[\widehat{E}_{\phi}(\bm{X})]\\
=& \sum_{\bm{X}}\widehat{E}_{\phi}(\bm{X})M(\bm{X})+(1-\widehat{E}_{\phi}(\bm{X}))(1-M(\bm{X})).\notag
\end{align}
Strictly speaking, \eqref{eq:cost1} is not equivalent to RPMD minimization. Maximizing $\log P[\widehat{E}_{\phi}]$ implies \emph{multiplying} $Pr[\widehat{E}_{\phi}(\bm{X})]$ over $\bm{X}\in\mathbb{Z}^2$. However, \eqref{eq:cost1} is equivalent to the $L^1$ minimization problem:
\begin{align}\label{eq:cost2}
    \phi_{opt2}=\arg \min_{\phi} \sum_{\bm{X}} \left|M(\bm{X})-\widehat{E}_{\phi}(\bm{X})\right|.
\end{align}
(Proof: Penalty for choosing $\widehat{E}_{\phi}(\bm{X})=1$ is $1-M(\bm{X})$ in \eqref{eq:cost2}, which is equivalent to reward of $M(\bm{X})$ in \eqref{eq:cost1}.)

Lastly, consider a minimization of classification error:
\begin{align}\label{eq:cost3}
    \phi_{opt3} = \arg\min_{\phi}\sum_{\bm{X}}\left|E_{opt}(\bm{X})-\widehat{E}_{\phi}(\bm{X})\right|,
\end{align}
where $E_{opt}$ is the theoretically optimal classifier defined in \eqref{eq:E_opt}. Given sufficient data, \eqref{eq:cost3} is statistically equivalent to \eqref{eq:cost1} and \eqref{eq:cost2}. Proof is provided in Appendix~\ref{sec:proof}.

\section{Application: Calibration} \label{sec:estCameraParams}

A key to calculating the event likelihood from APS and IMU data is knowing the log contrast sensitivity $\varepsilon$ in \eqref{eq:events}. In theory, this parameter value is controlled by registers in neuromorphic cameras programmed by the user. In practice, the programmed register values change the behavior of the DVS sensors, but the exact thresholding value remains unknown ~\cite{bryner2019event}. Similarly, the gain and offset values $a,b,\alpha,\beta$ in DVS \eqref{eq:J} and APS are not easily observable and are difficult to determine precisely. It is of concern because the offset value $O$ in \eqref{eq:V} is $O=\beta+\alpha b/a$ (see Appendix~\ref{sec:proof}). Offset allows mapping the small linear range of the APS to the large dynamic range of the DVS.

In this work, we make use of the EPM to calibrate the thresholding value $\varepsilon$ and the offset $O$ from the raw DVS data. Recalling \eqref{eq:EPM} and \eqref{eq:logPE}, the probabilistic quantities $Pr[H_0]$ and $\log Pr[E]$ are parameterized in part by $\varepsilon$ and $O$. Hence, rewriting them more precisely as $Pr[H_0|\varepsilon,O]$ and  $\log Pr[E|\varepsilon,O]$, respectively, we formulate a maximum likelihood estimate as follows:
\begin{align}\label{eq:mle}
    (\widehat{\varepsilon},\widehat{O})
    =\arg\max_{(\varepsilon,O)}\log P[E|\varepsilon,O],
\end{align}
where $E:\mathbb{Z}^2\to\{0,1\}$ denotes the event indicator for the unprocessed, noisy DVS data. Solution to \eqref{eq:mle} is not required for network inference. These estimated values are \emph{only} used to compute EPM for use in benchmarking (Section~\ref{sec:benchmark}) and denoising (Section~\ref{eventDenoising}).

\section{Experiments} \label{experiments}

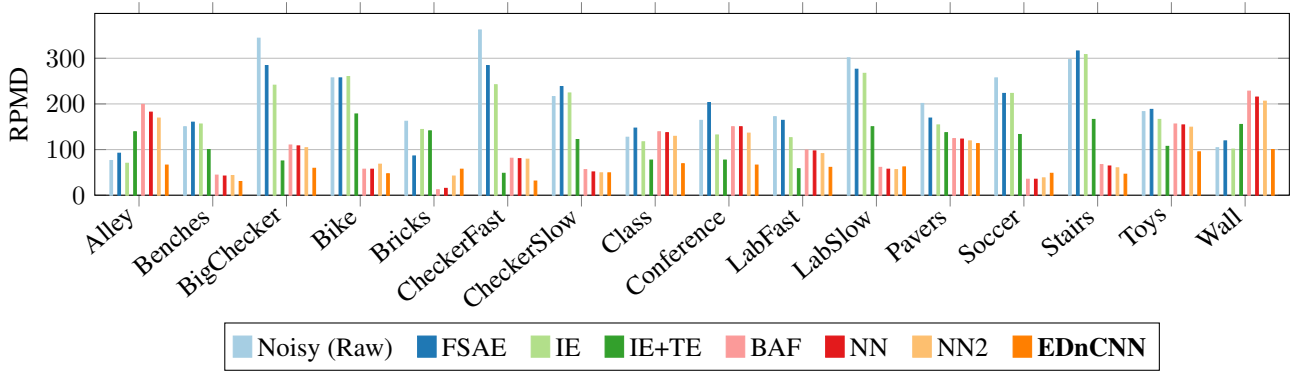
\begin{figure*}[ht]
\centering
\begin{tikzpicture}
\begin{axis}[
    ybar,
    cycle list name=Paired-10,
	bar width=1.0pt,
    enlarge x limits=0.04,
    ymin=0,
    legend style={at={(0.5,-0.70)},
      anchor=north,legend columns=-1,
      /tikz/every even column/.append style={column sep=0.25cm}}
      ,
    ylabel={RPMD},
    ymajorgrids,
    ytick={0, 100,200,300},
    symbolic x coords={Alley, Benches, BigChecker, Bike, Bricks, CheckerFast, CheckerSlow, Class, Conference, LabFast, LabSlow, Pavers, Soccer, Stairs, Toys, Wall},
    xtick=data,
    x tick label style={rotate=40,anchor=east},
    width=1.0\textwidth,
    height=4.0cm,
    every axis plot post/.append style={
            fill=.!100,
        }
    ]
\addplot+[ybar] plot coordinates {(Alley,76) (Benches,150) (BigChecker,344) (Bike,257) (Bricks,162) 
(CheckerFast,362) (CheckerSlow,216) (Class,127) (Conference,164) (LabFast,172) (LabSlow,301) (Pavers,201) 
(Soccer,257) (Stairs,297) (Toys,183) (Wall,104)};
\addplot+[ybar] plot coordinates {(Alley,92) (Benches,160) (BigChecker,284) (Bike,257) (Bricks,86) 
(CheckerFast,284) (CheckerSlow,238) (Class,147) (Conference,203) (LabFast,164) (LabSlow,276) (Pavers,169) 
(Soccer,223) (Stairs,316) (Toys,188) (Wall,119)};
\addplot+[ybar] plot coordinates {(Alley,70) (Benches,156) (BigChecker,241) (Bike,260) (Bricks,144) 
(CheckerFast,242) (CheckerSlow,224) (Class,117) (Conference,132) (LabFast,126) (LabSlow,267) (Pavers,154) 
(Soccer,223) (Stairs,308) (Toys,166) (Wall,101)};
\addplot+[ybar] plot coordinates {(Alley,139) (Benches,100) (BigChecker,75) (Bike,178) (Bricks,141) 
(CheckerFast,48) (CheckerSlow,122) (Class,77) (Conference,77) (LabFast,58) (LabSlow,150) (Pavers,137) 
(Soccer,133) (Stairs,166) (Toys,107) (Wall,155)};
\addplot+[ybar] plot coordinates {(Alley,199) (Benches,44) (BigChecker,110) (Bike,57) (Bricks,12) 
(CheckerFast,81) (CheckerSlow,56) (Class,139) (Conference,150) (LabFast,99) (LabSlow,61) (Pavers,124) 
(Soccer,35) (Stairs,67) (Toys,156) (Wall,228)};
\addplot+[ybar] plot coordinates {(Alley,182) (Benches,42) (BigChecker,108) (Bike,57) (Bricks,15) 
(CheckerFast,80) (CheckerSlow,51) (Class,137) (Conference,150) (LabFast,97) (LabSlow,57) (Pavers,123) 
(Soccer,35) (Stairs,64) (Toys,154) (Wall,215)};
\addplot+[ybar] plot coordinates {(Alley,169) (Benches,43) (BigChecker,104) (Bike,68) (Bricks,42) 
(CheckerFast,79) (CheckerSlow,49) (Class,129) (Conference,136) (LabFast,91) (LabSlow,56) (Pavers,119) 
(Soccer,38) (Stairs,60) (Toys,149) (Wall,206)};
\addplot+[ybar] plot coordinates {(Alley,66) (Benches,30) (BigChecker,59) (Bike,47) (Bricks,57) 
(CheckerFast,31) (CheckerSlow,49) (Class,69) (Conference,66) (LabFast,61) (LabSlow,62) (Pavers,113) 
(Soccer,48) (Stairs,46) (Toys,95) (Wall,100)};
\legend{\strut Noisy (Raw), \strut FSAE, \strut IE, \strut IE+TE, \strut BAF, \strut NN, \strut NN2, \strut \textbf{EDnCNN}}
\end{axis}
\end{tikzpicture}
\caption{Benchmark scores of denoising algorithms across 16 scenarios. Smaller RPMD values indicate better denoising performance.}
\label{fig:benchmark}
\end{figure*}

\subsection{Dataset: DVSNOISE20} \label{sec:data}

Data was collected using a DAVIS346 neuromorphic camera. It has a resolution of 346 $\times$ 260 pixels, dynamic range of 56.7 and 120dB for APS and DVS respectively, a latency of 20$\mu$s, and a 6-axis IMU. As discussed in Section~\ref{sec:limitations}, Theorem~\ref{thm:Mx} is valid in absence of translational camera motion and moving objects. Movement of the camera was restricted by a gimbal (Figure~\ref{fig:gimbal}), and the IMU was calibrated before each collection. Only stationary scenes were selected, avoiding saturation and severe noise in the APS. 

%We did not manually remove a handful of hot pixels in DVS, generating events continuously regardless of the scene content. However, most denoising methods correctly identified these pixels as noisy. 

The APS framerate (41-56 fps; $\eta$ range 17 to 24 ms) was maximized using a fixed exposure time ($\tau$ range 0.13ms to 6ms) per scene. Since EPM labeling is only valid during the APS exposure time, benchmarking (Section~\ref{sec:benchmark}) and training EDnCNN (Section~\ref{sec:training}) are restricted to events occurring during this time. However, a large data volume can be acquired easily by extending the length of collection. In addition, we calibrated APS fixed pattern noise and compensated for the spatial non-uniformity of pixel gain and bias. 

We obtained 16 indoor and outdoor scenes of noisy, real-world data to form \emph{DVSNOISE20}. Examples are shown in Figure~\ref{tab:dataset_EPM}. Each scene was captured three times for $\approx$16 seconds, giving 48 total sequences with a wide range of motions. The calibration procedure outlined in Section~\ref{sec:estCameraParams} was completed for each sequence. The estimates of internal camera parameters were repeatable and had mean/standard deviation ratios of 21.44 ($O$), 13.58 ($\varepsilon_{pos}$), and 13.27 ($\varepsilon_{neg}$). The DVSNOISE20 dataset, calibration, and code are available at: \url{http://issl.udayton.edu}.

\subsection{Results}

To ensure fair evaluation, EDnCNN was trained using a leave-one-scene-out strategy for DVSNOISE20. In Figure~\ref{fig:benchmark}, the performance of EDnCNN is assessed by RPMD. EDnCNN improved the RPMD performance of the noisy data by an average gain of 148 points. Improvement from denoising was significant in all scenes except for the Alley and Wall sequences. These scenes have highly textured scene contents, which are a known challenge for neuromorphic cameras, and thus represent the worst case performance we expect from any denoising method. The RPMD performance of EDnCNN in these two scenes was no better or worse than noisy input.

\begin{figure*}[t]
\centering
\begin{tabular}{m{0.01\textwidth}m{0.165\textwidth}m{0.165\textwidth}m{0.165\textwidth}m{0.165\textwidth}m{0.165\textwidth}}
\rotatebox{90}{IE+TE} & \includegraphics[width=1.07\linewidth]{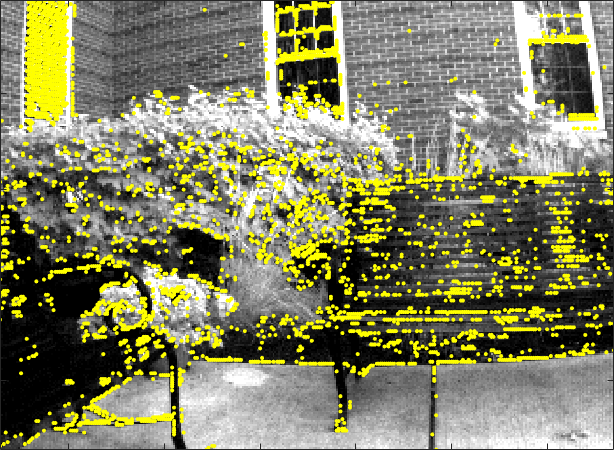} & \includegraphics[width=1.07\linewidth]{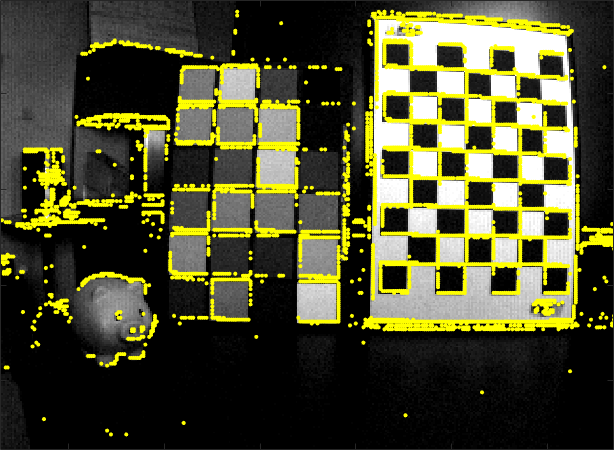} & \includegraphics[width=1.07\linewidth]{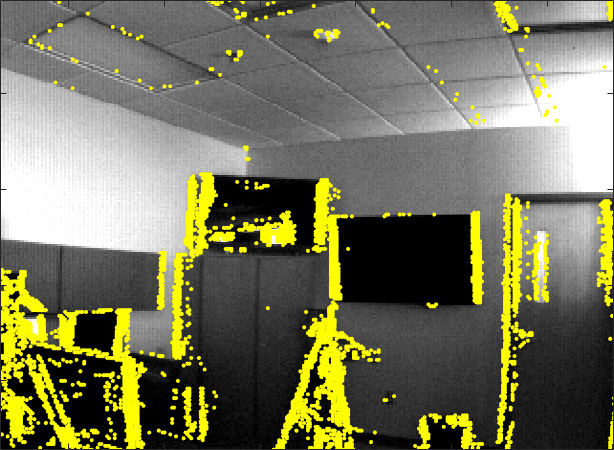} & 
\includegraphics[width=1.07\linewidth]{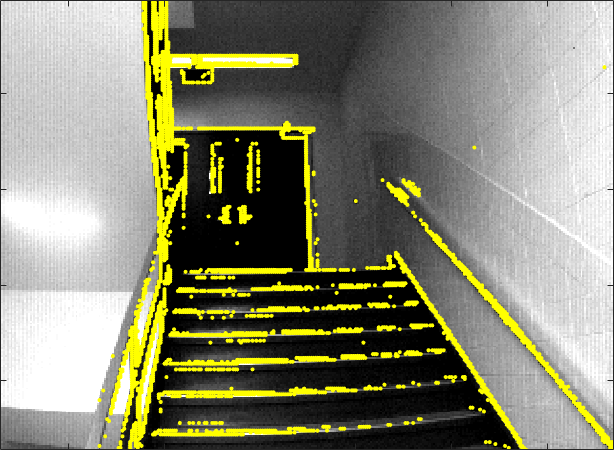} & \includegraphics[width=1.07\linewidth]{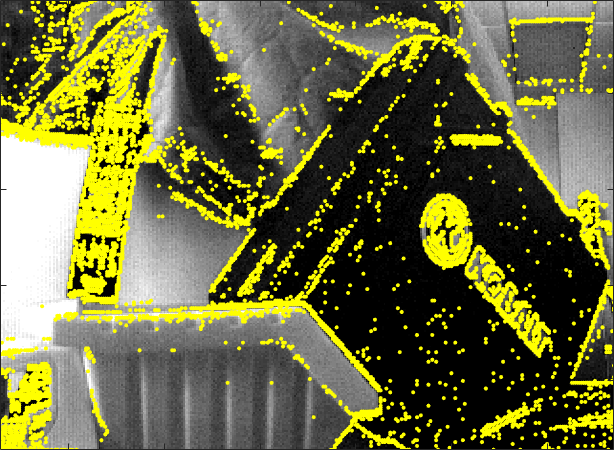} \\
\rotatebox{90}{BAF} & \includegraphics[width=1.07\linewidth]{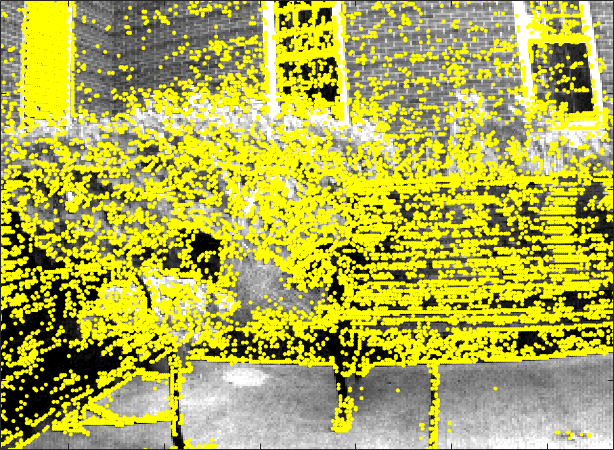} & \includegraphics[width=1.07\linewidth]{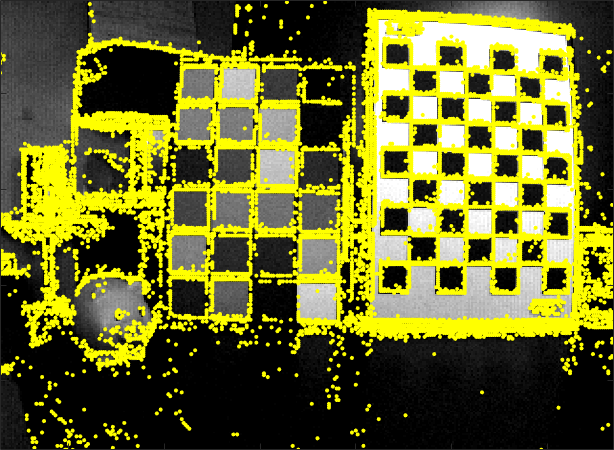} & \includegraphics[width=1.07\linewidth]{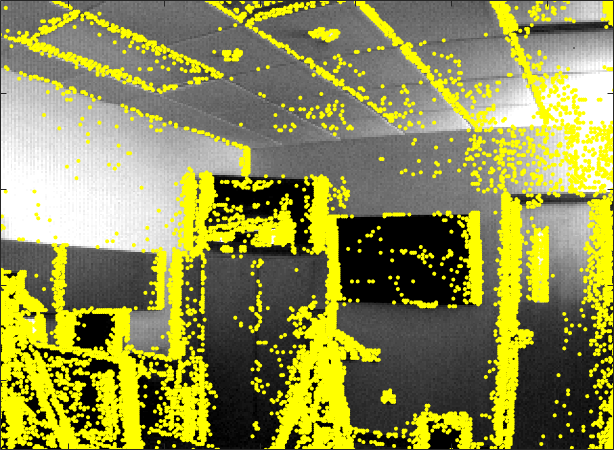} & 
\includegraphics[width=1.07\linewidth]{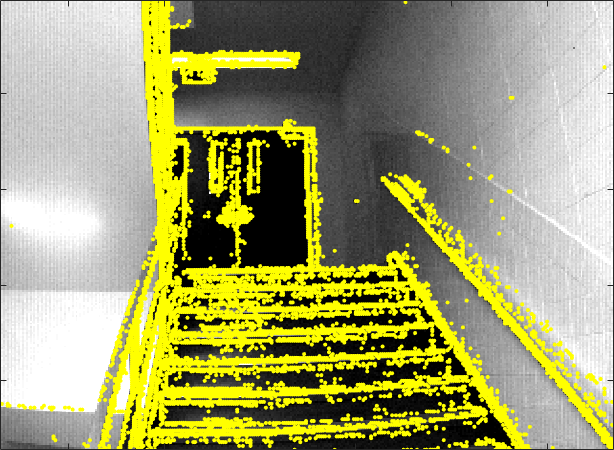} & \includegraphics[width=1.07\linewidth]{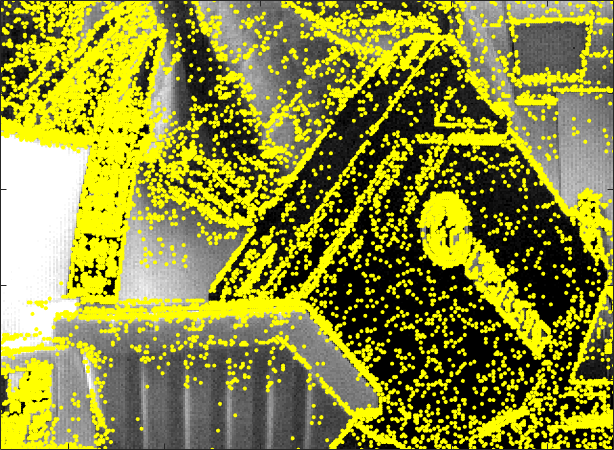} \\
\rotatebox{90}{NN2} & \includegraphics[width=1.07\linewidth]{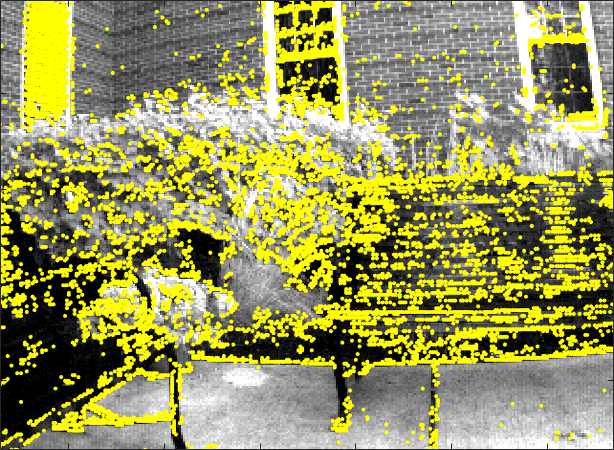} & \includegraphics[width=1.07\linewidth]{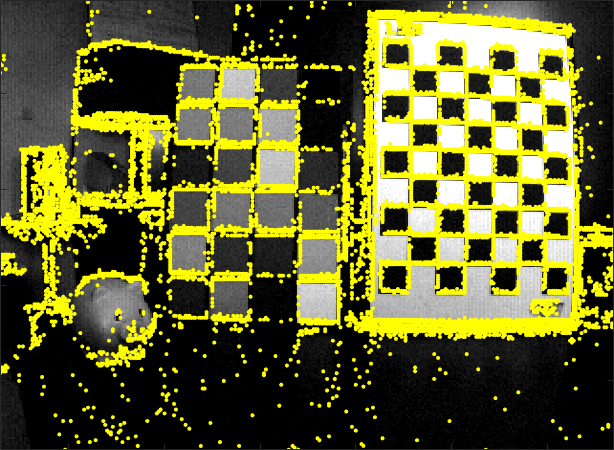} & \includegraphics[width=1.07\linewidth]{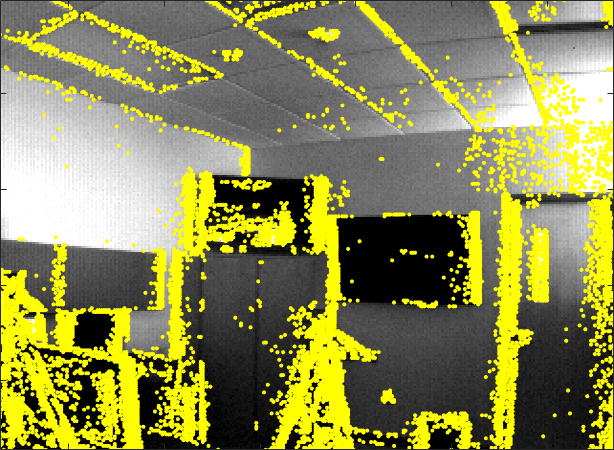} & 
\includegraphics[width=1.07\linewidth]{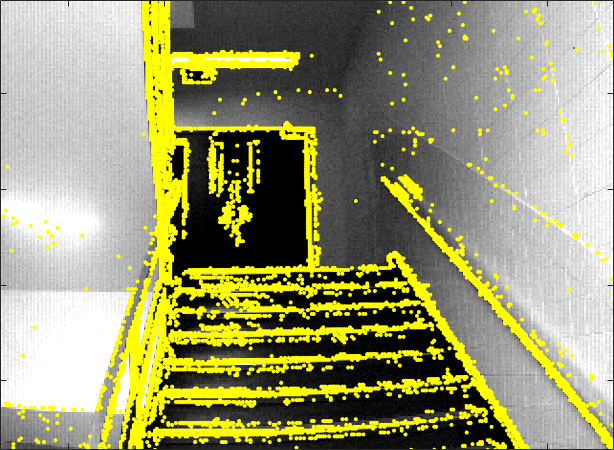} & \includegraphics[width=1.07\linewidth]{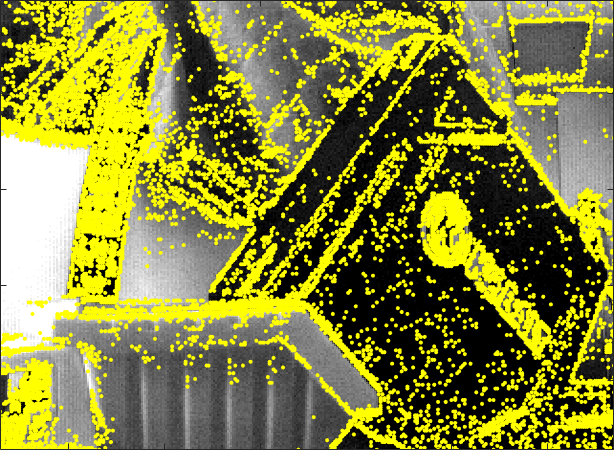} \\
\rotatebox{90}{\textbf{EDnCNN}} &  \includegraphics[width=1.07\linewidth]{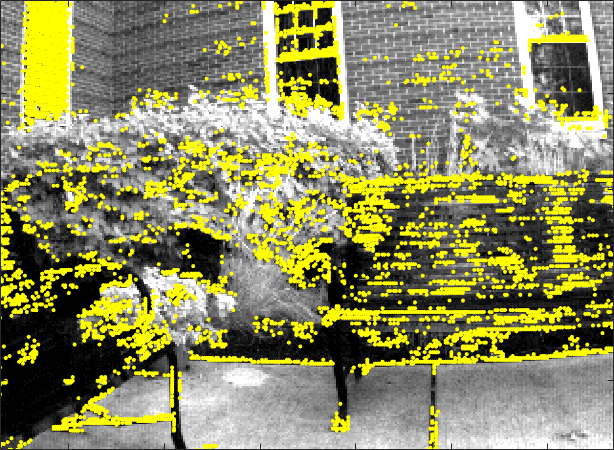} & \includegraphics[width=1.07\linewidth]{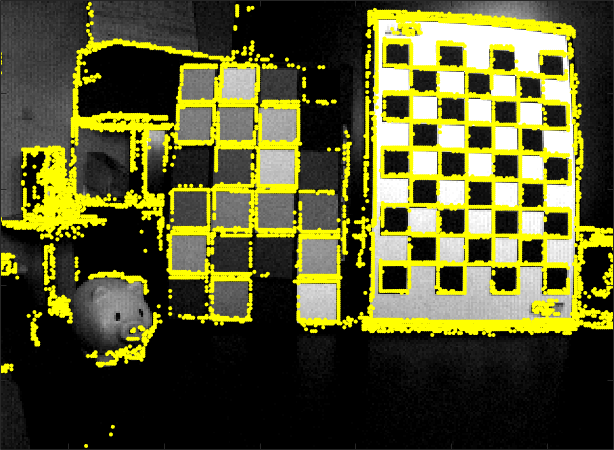} & \includegraphics[width=1.07\linewidth]{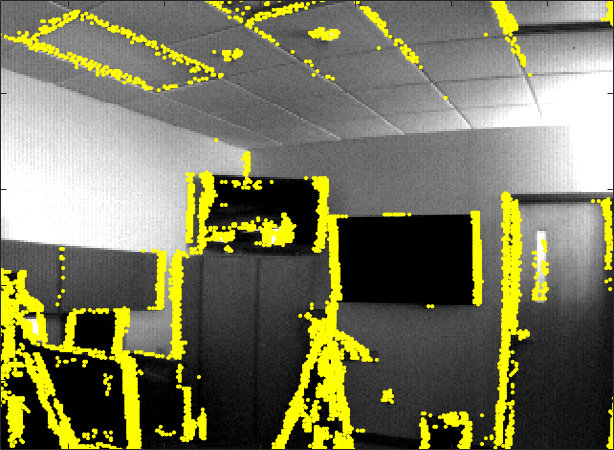} & 
\includegraphics[width=1.07\linewidth]{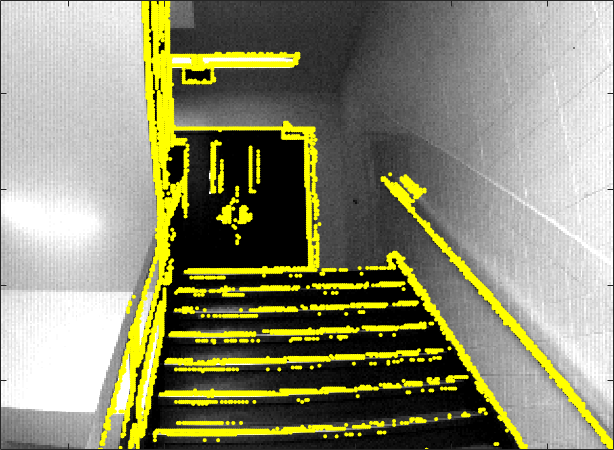} & \includegraphics[width=1.07\linewidth]{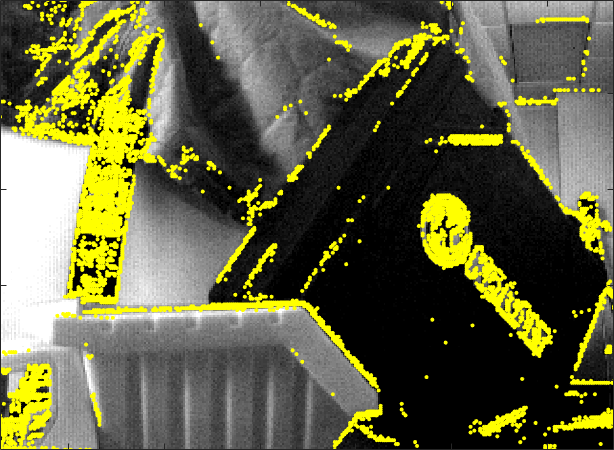} \\
\multicolumn{1}{r}{} & \multicolumn{1}{c}{Benches} & \multicolumn{1}{c}{CheckerSlow} & \multicolumn{1}{c}{LabFast} & \multicolumn{1}{c}{Stairs} & \multicolumn{1}{c}{Toys}
\end{tabular}%
\caption{DVSNOISE20 denoising results from four different algorithms. Denoised DVS events (yellow) overlaid on APS frames.}
\label{tab:imageCompareTable}
\end{figure*}

EDnCNN was benchmarked against other state-of-the-art denoising methods: filtered surface of active events (FSAE~\cite{mueggler2017fast}), inceptive and trailing events (IE \& IE+TE~\cite{baldwin2019inceptive}), background activity filter (BAF~\cite{delbruck2008frame}), and nearest neighbor (NN \& NN2~\cite{padala2018noise}). RPMD scores are reported in Figure~\ref{fig:benchmark} for each scene. FSAE and IE did not improve significantly over noisy input, but did reduce total data volume. IE+TE improved RPMD performance while reducing data volume and generated a top score in the LabFast sequence. BAF, NN, and NN2 work and behave similarly, outperforming EDnCNN in 3 of 16 scenes. However, the performances of these methods were significantly more sensitive than EDnCNN. EDnCNN outperformed other denoising methods in 12 of 16 scenes, with a statistically significant p-value of $0.00248$ via the Wilcoxon signed-rank test.

Figure~\ref{tab:imageCompareTable} shows examples of denoised DVS events (superimposed on APS image for visualization). Qualitatively, IE+TE, BAF, and NN2 pass events that are spatially isolated, making it more difficult to distinguish edge shapes compared to EDnCNN. EDnCNN removes events that do not correspond to edges, and enforces a strong agreement with the EPM label as designed (see Figure~\ref{tab:dataset_EPM}).

%EDnCNN can underdetect events in regions with low pixel contrast, but events that EDnCNN passed were nearly all included in $E_{opt}$ (i.e.~$E_{opt}(\bm{X})=1$ when $\widehat{E}_{\phi}(\bm{X})=1$).

%\newcommand{\STAB}[1]{\begin{tabular}{@{}c@{}}#1\end{tabular}}
%{cccccc}

\begin{figure}[t]
    \centering
    \includegraphics[width=1.0\linewidth]{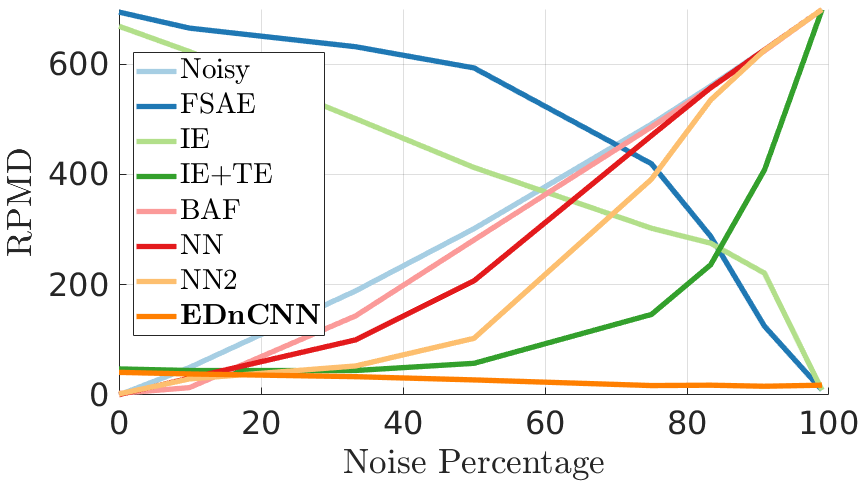}
   \caption{Simulated Results: Random noise was injected into simulated data to test each algorithm's robustness to input noise levels.}
\label{fig:simNoise}
\end{figure}

\begin{figure}[h]
\centering
%\begin{center}
\begin{tabular}{m{0.01\textwidth}m{0.17\textwidth}m{0.17\textwidth}}
\rotatebox{90}{Noisy} & \includegraphics[width=1.07\linewidth]{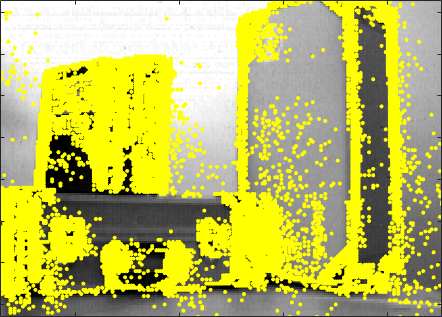} & 
\includegraphics[width=1.07\linewidth]{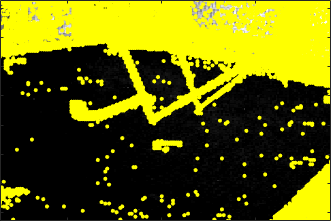} \\
\rotatebox{90}{\textbf{EDnCNN}} & \includegraphics[width=1.07\linewidth]{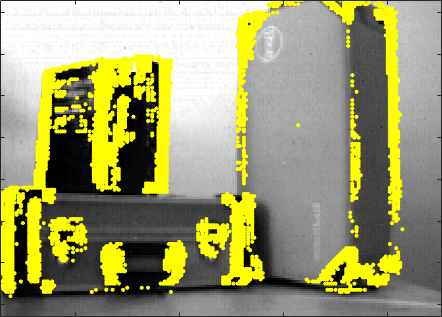} & 
\includegraphics[width=1.07\linewidth]{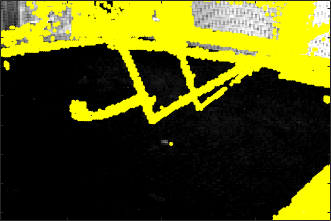} \\
\multicolumn{1}{r}{} & \multicolumn{1}{c}{(a) DVSFLOW16~\cite{rueckauer2016evaluation}} & \multicolumn{1}{c}{(b) IROS18~\cite{mitrokhin2018event}}
\end{tabular}%
\caption{EDnCNN results on published datasets. (a) DVS Optical Flow dataset with non-rotational camera motion~\cite{rueckauer2016evaluation}. (b) Extreme Event Dataset (EED) with multiple object motions~ \cite{mitrokhin2018event}.
}
\label{fig:moreExamples}
%\end{center}
\end{figure}

\subsection{Robustness to Assumptions and Dataset} \label{sec:otherexp}

To test for robustness, we also benchmarked on simulated DVS data from ESIM~\cite{rebecq2018esim}. ESIM is an event camera simulator allowing user-specified 3D scenes, lighting, and camera motion. In our experiments, we interpret DVS data simulated from a virtual scene as an output from a noise-free neuromorphic camera. We then injected additional random events (i.e.~BA noise) into the scene. Figure~\ref{fig:simNoise} shows the result of RPMD benchmarking on synthetic data as a function of BA noise percentage. As expected, the RPMD performance of noisy data scales linearly as the percentage of noise increases. Since EDnCNN was not trained against noise-free data, it has a slightly lower performance than some other methods at 0\% BA noise level. By contrast, IE+TE, BAF, NN, and NN2 underdetect noise as event count increases, deteriorating performance.

%However, at moderate and high noise percentage, the \todo{density-based} algorithms failed while EDnCNN performance remained robust.

Figure~\ref{fig:moreExamples} shows example sequences with non-rotational camera motion~\cite{rueckauer2016evaluation} and multiple moving objects~\cite{mitrokhin2018event}. Qualitatively, EDnCNN denoising seems consistent with stationary scene performance in Figure~\ref{tab:imageCompareTable}. Additional analysis on other datasets and further examples may be found in supplementary material and Appendix~\ref{sec:proof}.

\section{Conclusion} \label{conclusion}

Contrast sensitivity in neuromorphic cameras is primarily limited by noise. In this paper, we present five major contributions to address this issue. We rigorously derived a method to assign a probability (EPM) for observing events within a very short time interval. We proposed a new benchmarking metric (RPMD) that enables quantitative comparison of denoising performance on real-world neuromorphic cameras. We developed EDnCNN, a neural network-based event denoising method trained to minimize RPMD. We showed that internal camera parameters can be estimated based on natural scene output. We collected a new benchmarking dataset for denoising along with the EPM labeling tools (DVSNOISE20). Quantitative and qualitative assessment verified that EDnCNN outperforms prior art. EDnCNN admits higher contrast sensitivity (i.e.~detection of scene content obscured by noise) and would vastly enhance neuromorphic vision across a wide variety of applications.

%\clearpage

{\small
\bibliographystyle{ieee_fullname}
\bibliography{egbib}
}

\clearpage

\appendix

\section{Appendix}
\label{sec:proof}

\subsection{Proof of Theorem 1}

To simplify notation, we omit below pixel location $\bm{X}\in\mathbb{Z}^2$ from events $\{t_i,p_i\}$ whenever it is unambiguous from the context. In a noise-free neuromorphic camera hardware, thresholding in \eqref{eq:events} imply equality at threshold:
\begin{align}\label{eq:j_diff}
    |J(\bm{X},t_i)-J(\bm{X},t_{i-1})| = \varepsilon.
\end{align}
Substituting a Taylor series expansion of the form
\begin{align}\label{eq:Jtswap}
    J(\bm{X},t_{i-1}) \approx J(\bm{X},t_i) + (t_i-t_{i-1}) J_t(\bm{X},t_i),
\end{align}
where $J_t(\bm{X},t)=\frac{\partial}{\partial t}J(\bm{X},t)$, the relation in \eqref{eq:j_diff} may be rewritten as:
\begin{align}
\begin{split}
&|J(\bm{X},t_i)-(J(\bm{X},t_i) + (t_i-t_{i-1}) J_t(\bm{X},t_i))|\\
&=|(t_i-t_{i-1})J_t(\bm{X},t_i)|= \varepsilon.
\end{split}
\end{align}
Or equivalently,
\begin{align}\label{eq:ratio}
t_i-t_{i-1}= \frac{\varepsilon}{|J_t(\bm{X},t_i)|}.
\end{align}
In other words, $\frac{\varepsilon}{|J_t(\bm{X},t_i)|}$ is the ``rate'' at which events are generated. Hence the probability that an event falls within a time interval $[t,t+\tau)$ is 
\begin{align}
    M(\bm{X})
    =\begin{cases}
    \tau\frac{|J_t(\bm{X},t)|}{\varepsilon}&\text{if $\tau<\frac{\epsilon}{|J_t(\bm{X},t)|}$}\\
    1&\text{else}.
    \end{cases}
\end{align}
Intuition here is that if the rate $t_i-t_{i-1}$ is smaller than the window size $\tau$, the event $(t_i,p_i)$ and/or $(t_{i-1},p_{i-1})$ will have taken place within this time interval. However, if the rate $t_i-t_{i-1}$ is larger than the window size $\tau$, then events do not necessarily occur within this time interval. As is obvious from Figure~\ref{fig:eventGen}, $M(\bm{X})$ scales proportionally to the time window $\tau$ and inverse proportionally to the rate $t_i-t_{i-1}$. 

To compute $J_t(\bm{X},t)$ from APS and IMU, we draw on the well established principles of optical flow. Known as ``brightness constancy constraint,'' spatial translation of pixels over time obeys the following rule~\cite{horn1981determining}:
\begin{equation}\label{eq:constancy}
 J(\bm{X}+\Delta\bm{X},t+\Delta t) = J(\bm{X},t),
\end{equation}
where $\Delta\bm{X}=(\Delta x,\Delta y)$ refers to the pixel translation occurring during the time interval $\Delta t$. By Taylor expansion, we obtain the classical ``optical flow equation'':
\begin{align} \label{eq:OFeq}
 J_t(\bm{X},t) \approx -\nabla J(\bm{X},t)\bm{V}(\bm{X},t),
  \end{align} 
where 
\begin{align}
\begin{split}
    \nabla J(\bm{X},t)  =&(J_x(\bm{X},t),J_y(\bm{X},t))\\
    :=&\left(\frac{\partial}{\partial x}J(\bm{X},t),\frac{\partial}{\partial y}J(\bm{X},t)\right)
\end{split}\\
\begin{split}
    \bm{V}(\bm{X},t)=&(v_x,v_y):=\left(\frac{\Delta x}{\Delta t}\frac{\Delta y}{\Delta t}\right)^T
\end{split}
\end{align} 
denotes the spatial gradient and the flow field vector of log intensity $J:\mathbb{Z}^2\to\mathbb{R}$ at pixel $\bm{X}\in\mathbb{Z}^2$ at time $t\in\mathbb{R}$. We obtain the pixel velocity $\bm{V}(\bm{X},t)$ from the IMU; and spatial gradient $\nabla J(\bm{X},t)$ from APS.

\begin{figure*}[t]
\centering
\begin{tabular}{m{0.01\textwidth}m{0.165\textwidth}m{0.165\textwidth}m{0.165\textwidth}m{0.165\textwidth}m{0.165\textwidth}}
% \rotatebox{90}{Image} & \includegraphics[width=1.07\linewidth]{pics/of/1_image.png} & \includegraphics[width=1.07\linewidth]{pics/of/2_image.png} & \includegraphics[width=1.07\linewidth]{pics/iros/1_image.png} & 
% \includegraphics[width=1.07\linewidth]{pics/iros/5_image.png} & \includegraphics[width=1.07\linewidth]{pics/iros/6_image.png} \\
\rotatebox{90}{Noisy} & \includegraphics[width=1.07\linewidth]{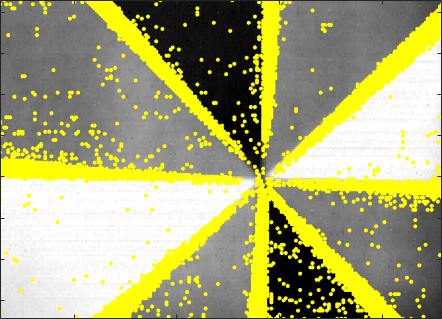} & \includegraphics[width=1.07\linewidth]{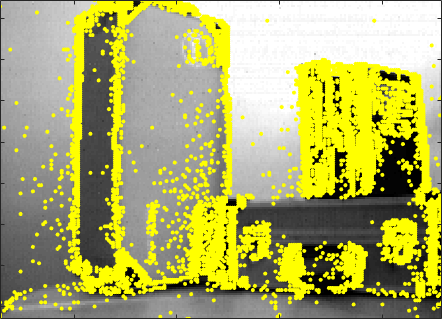} & \includegraphics[width=1.07\linewidth]{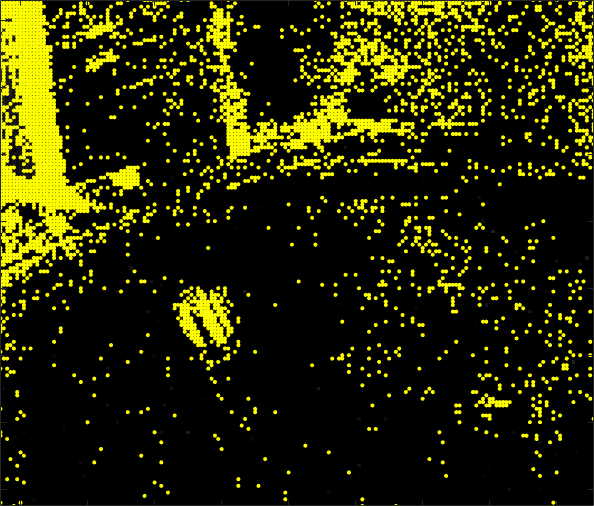} & 
\includegraphics[width=1.07\linewidth]{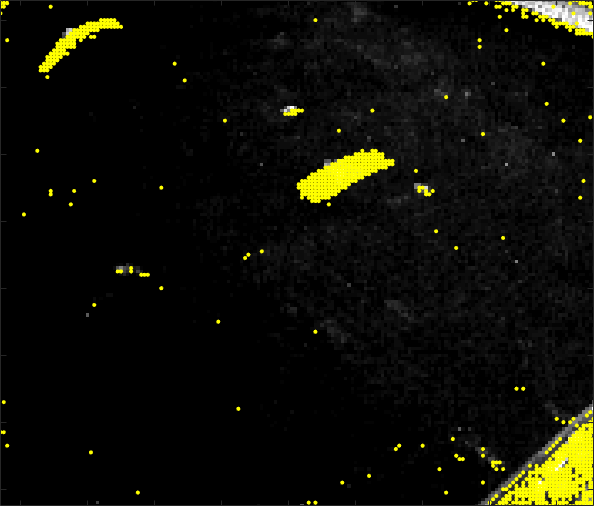} & \includegraphics[width=1.07\linewidth]{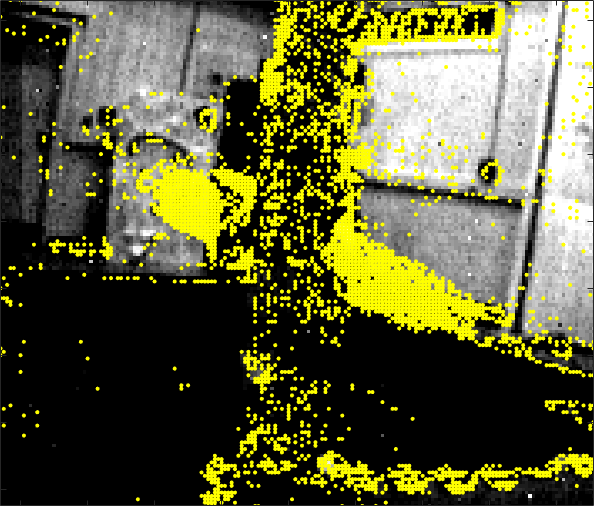} \\
\rotatebox{90}{IE+TE} & \includegraphics[width=1.07\linewidth]{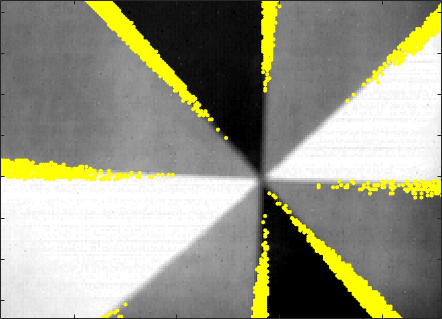} & \includegraphics[width=1.07\linewidth]{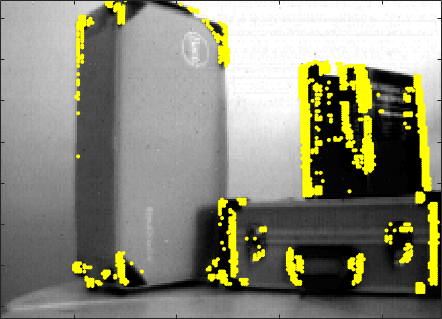} & \includegraphics[width=1.07\linewidth]{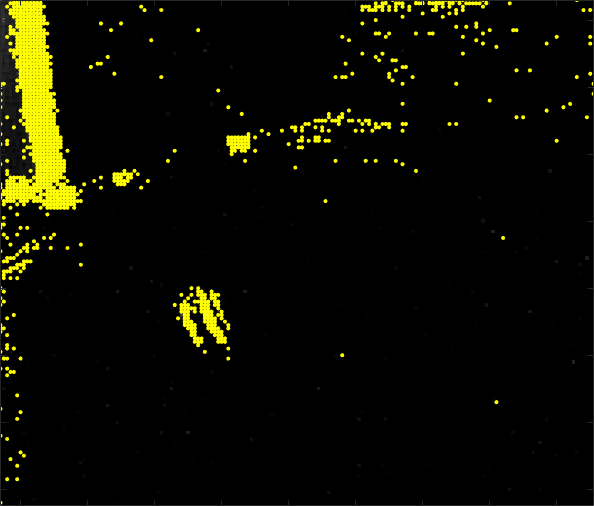} & 
\includegraphics[width=1.07\linewidth]{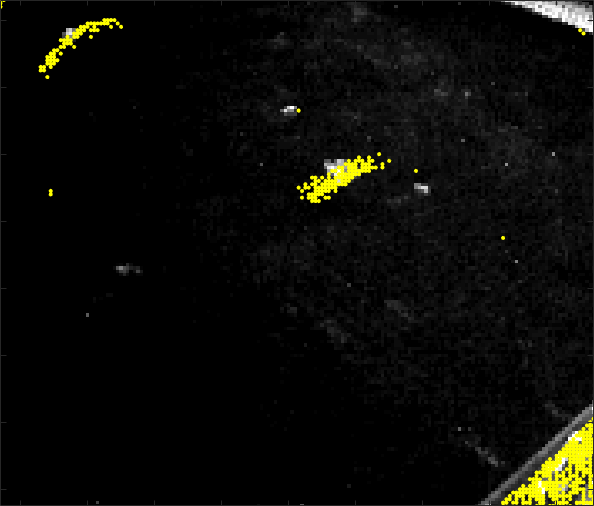} & \includegraphics[width=1.07\linewidth]{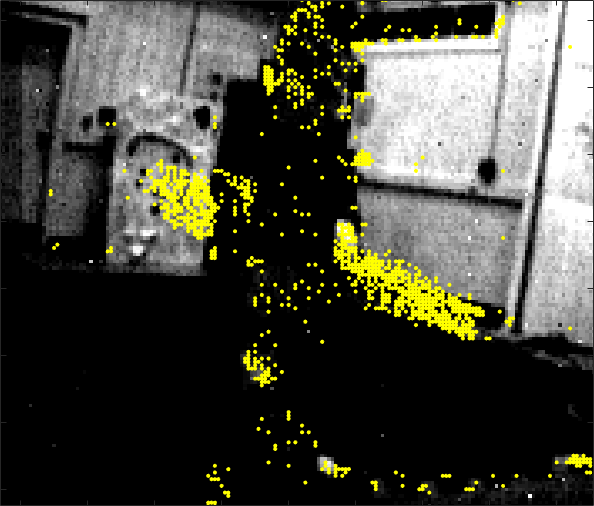} \\
\rotatebox{90}{BAF} & \includegraphics[width=1.07\linewidth]{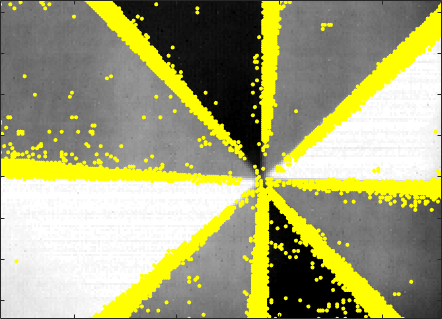} & \includegraphics[width=1.07\linewidth]{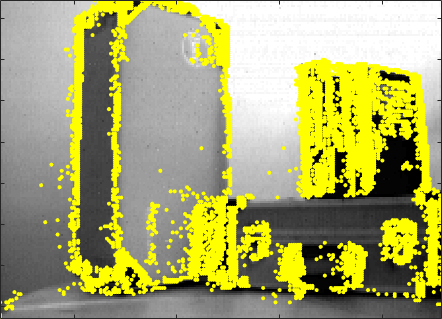} & \includegraphics[width=1.07\linewidth]{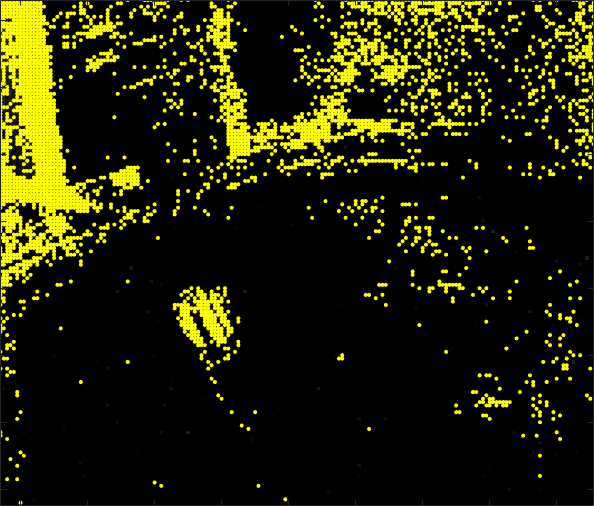} & 
\includegraphics[width=1.07\linewidth]{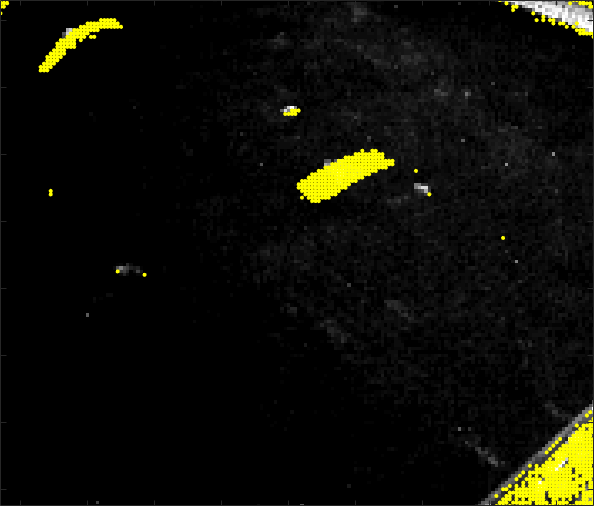} & \includegraphics[width=1.07\linewidth]{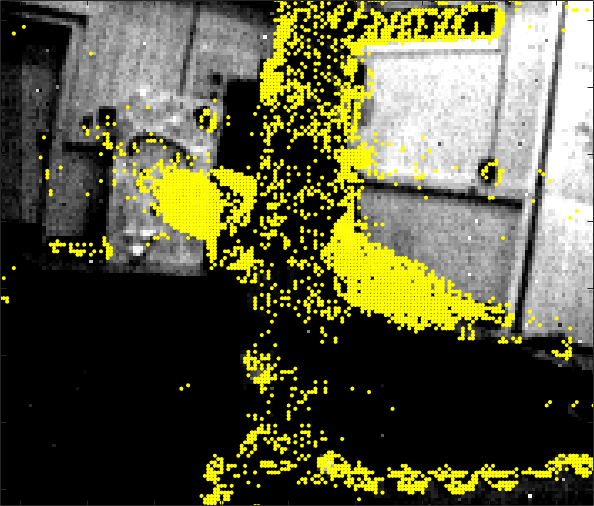} \\
\rotatebox{90}{NN2} & \includegraphics[width=1.07\linewidth]{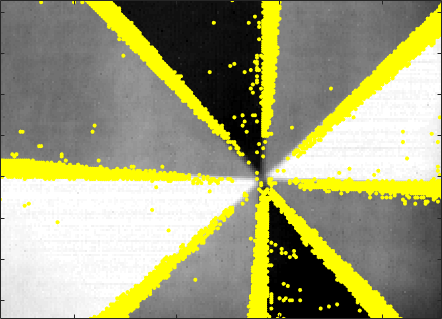} & \includegraphics[width=1.07\linewidth]{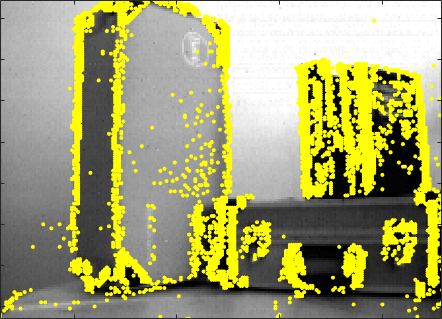} & \includegraphics[width=1.07\linewidth]{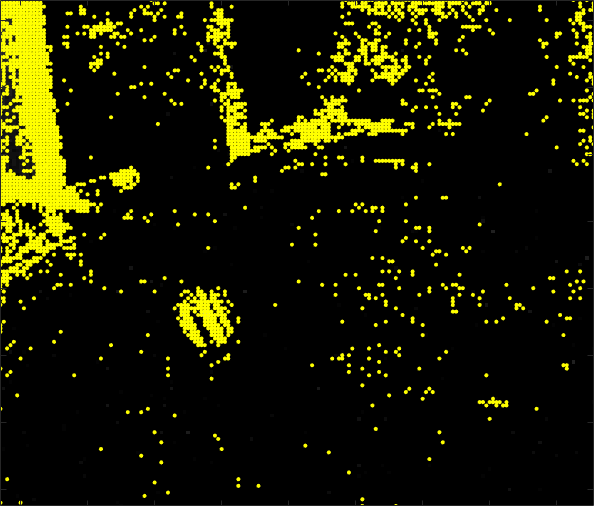} & 
\includegraphics[width=1.07\linewidth]{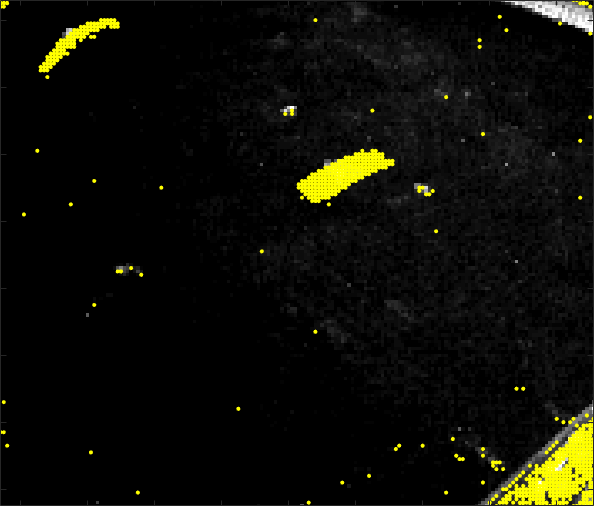} & \includegraphics[width=1.07\linewidth]{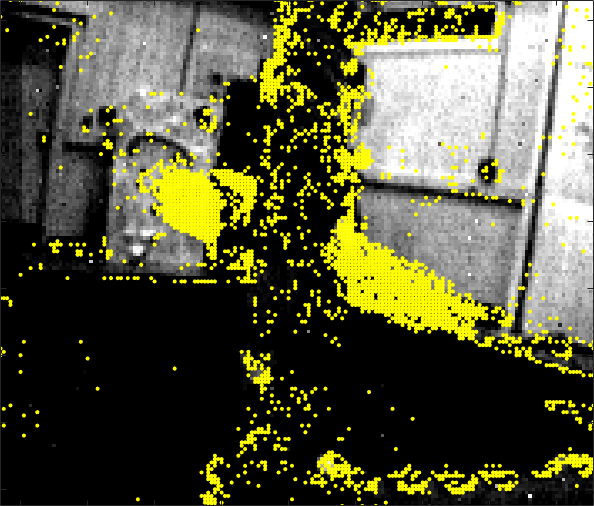} \\
\rotatebox{90}{\textbf{EDnCNN}} &  \includegraphics[width=1.07\linewidth]{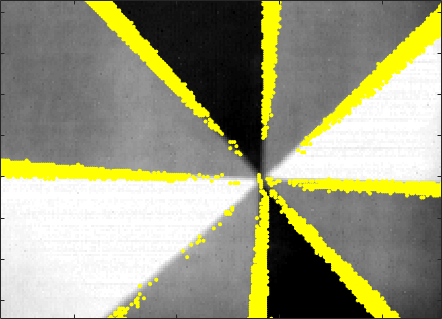} & \includegraphics[width=1.07\linewidth]{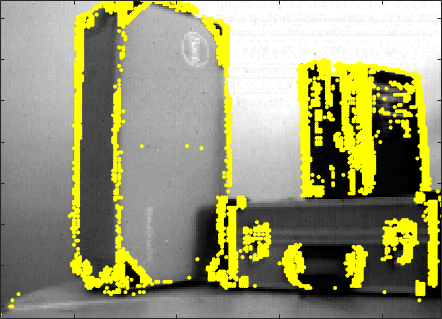} & \includegraphics[width=1.07\linewidth]{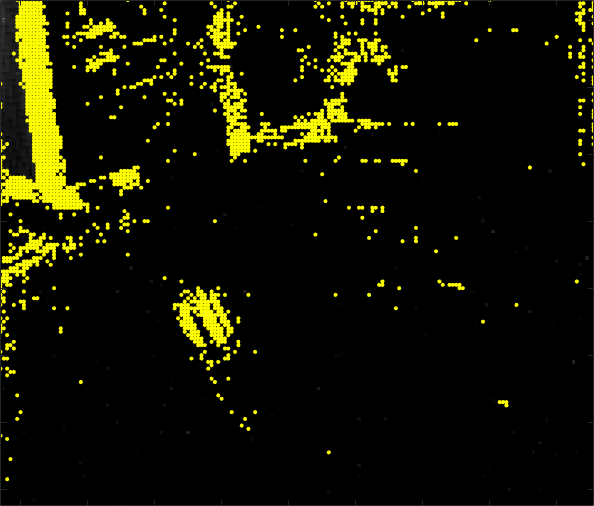} & 
\includegraphics[width=1.07\linewidth]{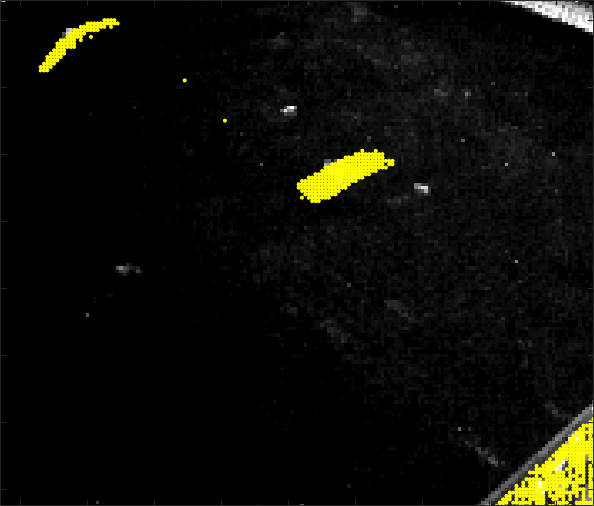} & \includegraphics[width=1.07\linewidth]{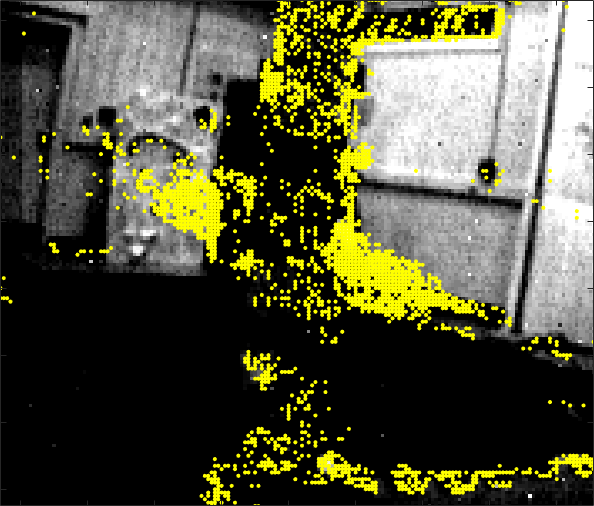} \\
\multicolumn{1}{r}{} & \multicolumn{1}{c}{\begin{tabular}[b]{@{}c@{}}Disk\\DVSFLOW16~\cite{rueckauer2016evaluation}\end{tabular}} & \multicolumn{1}{c}{\begin{tabular}[b]{@{}c@{}}Boxes\\DVSFLOW16\end{tabular}} & \multicolumn{1}{c}{\begin{tabular}[b]{@{}c@{}}Fast Drone\\IROS18~\cite{mitrokhin2018event}\end{tabular}} & \multicolumn{1}{c}{\begin{tabular}[b]{@{}c@{}}3 Objects\\IROS18\end{tabular}} & \multicolumn{1}{c}{\begin{tabular}[b]{@{}c@{}}Occlusions\\IROS18\end{tabular}}
\end{tabular}%
\caption{Additional qualitative results from DVS Optical Flow and IROS18. (First Row) A single APS frame from each dataset. The remaining rows show the APS frame overlayed with denoised DVS events from each algorithm. The APS images for columns "Fast Drone" and "3 Objects" have been contrast enhanced but remain dark due to limited signal. Limited APS signal does not impact DVS event generation.}

\label{tab:additionalImageCompareTable}
\end{figure*}

Consider the camera configuration in Figure~\ref{fig:gimbal}, where a camera on a rotational gimbal is observing a stationary scene. Let  $\bm{\theta}(t)=(\theta_x(t),\theta_y(t),\theta_z(t))^T$ represent the instantaneous 3-axis angular velocity of camera measured by IMU's gyroscope. Then the instantaneous pixel velocity $\bm{V}(\bm{X},t)$ stemming from \emph{yaw}, \emph{pitch}, and \emph{roll} rotations of the camera is computable as
\begin{align}\label{eq:pix_velocity1}
 \begin{pmatrix}v_x\\v_y\\0\end{pmatrix}
    =\bm{K}\begin{pmatrix}
    0&-\theta_z(t)&\theta_y(t)\\
    \theta_z(t)&0&-\theta_x(t)\\
    -\theta_y(t)&\theta_x(t)&0
    \end{pmatrix}\bm{K}^{-1}\begin{pmatrix}x\\y\\1\end{pmatrix},
\end{align}
where the camera intrinsic matrix 
\begin{align}
    \bm{K}=\begin{pmatrix}
    f&\kappa&c_x\\0&f&c_y\\0&0&1
    \end{pmatrix}
\end{align}
is characterized by focal length $f$, principal point $c_x,c_y$, and skew parameter $\kappa$ ($\kappa=0$ when the image sensor pixels are square). Hence, the pixel velocity is now entirely determined by the angular velocity and is decoupled from the scene content.

On the other hand, let  $A:\mathbb{Z}^2\times\mathbb{Z}\to\mathbb{R}$ be a synchronous APS output. APS makes measurements on  intensity video $I:\mathbb{Z}^2\times \mathbb{R}\to\mathbb{R}$ as follows:
\begin{align}\label{eq:I}
    A(\bm{X},k)=\alpha \int_{k\eta}^{k\eta+\tau} I(\bm{X},t) dt+\beta
\end{align}
where $k\in\mathbb{Z}$ denotes the frame number; $1/\eta$ is the frame rate; and $\alpha$ and $\beta$ are gain and black offset, respectively. The APS exposure time is denoted $\tau<\eta$ hereto also correspond to the time window $\tau$ hypothesis in \eqref{eq:hypothesis2} and EPM in \eqref{eq:EPM}. In absence of pixel motion, substituting \eqref{eq:I} into \eqref{eq:J}, $I:\mathbb{Z}^2\times\mathbb{Z}\to\mathbb{R}$ and $J:\mathbb{Z}\times \mathbb{R}\to \mathbb{R}$ yields the relationship
\begin{align}\label{eq:J2}
J(\bm{X},t)=\log\left(\frac{a}{\alpha\tau} A(\bm{X},t)-\frac{a\beta}{\alpha\tau}+b\right)
\end{align}
at time $t=k\eta$. Taking its spatial gradient yields
\begin{align}\label{eq:gradient1}
\begin{split}
    \nabla J(\bm{X},t) =& \frac{\nabla A(\bm{X})}{ A(\bm{X},t)-O}\\
O=&\beta+\alpha\tau b/a.
\end{split}
\end{align}
In presence of pixel motion, however, APS image $A:\mathbb{Z}^2\times\mathbb{Z}\to\mathbb{R}$ in \eqref{eq:I} is blurred. Assuming constant velocity $\bm{V}(\bm{X},t)=\bm{V}$ within time $t\in[k\eta,k\eta+\tau)$, the spatial gradient $\nabla A(\bm{X})$ is attenuated by $\tau|\bm{V}(\bm{X},t)|$ (proof below). Therefore, to correct for the attenuation we revise \eqref{eq:gradient1} as follows:
\begin{align}\label{eq:gradient1}
    \nabla J(\bm{X},t) =& \frac{\tau\nabla A(\bm{X})}{ A(\bm{X},t)-O}
    \begin{pmatrix}
    |V_x(\bm{X},t)|&0\\0&|V_y(\bm{X},t)|
    \end{pmatrix}.
\end{align}

To understand the impact of the blur on derivatives, consider a canonical edge image $I(\bm{X},t)=U(\bm{X}+\bm{V}t)$ (where $U$ is a unit step function in $x$ direction) crossing pixel $\bm{X}={0\choose 0}$ at time $t=0$ and frame $k=0$: 
\begin{align}
I(\bm{X},t)=U(\bm{X}+t\bm{V}(\bm{X},t)).
\end{align}
In absence of motion, the APS spatial derivative $A_x(\bm{X},t):=\frac{\partial}{\partial x}A(\bm{X},t)$ has the following value at pixel location $\bm{X}={0\choose 0}$ and frame $k=0$:
\begin{align}\label{eq:U_stationary}
    \begin{split}
    A_x(0,0,0)=&\alpha \int_{0}^{\tau} \left.\frac{\partial}{\partial x}U(\bm{X})\right|_{\bm{X}=(0,0)} dt+\beta\\
    =&\alpha \tau +\beta.
    \end{split}
\end{align}
By contrast, $A_x(\bm{X},0)$ with non-trivial motion $\bm{V}$ has the following form:
\begin{align}
\begin{split}
    &A_x(\bm{X},k)=
    \alpha \int_{k\eta}^{k\eta+\tau} \left.\frac{\partial}{\partial x}U(\bm{X})\right|_{\bm{X}=(0,0)} dt+\beta\\
    &=\alpha \int_{k\eta}^{k\eta+\tau} \delta(\bm{X}+t\bm{V}(\bm{X},t))(1+tV_x(\bm{X},t)) dt+\beta.
\end{split}
\end{align}
At $k=0$ and $\bm{X}={0\choose 0}$,
\begin{align}\label{eq:U_moving}
\begin{split}
    &A_x(0,0,0)\\
    &=\alpha \int_{0}^{\tau} \delta((0,0)+t\bm{V}(0,0,t))(1+tV_x(0,0,t)) dt+\beta\\
    &=\frac{\alpha}{|\bm{V}|}+\beta.
\end{split}
\end{align}
Comparing \eqref{eq:U_stationary} and \eqref{eq:U_moving}, we confirm that the spatial derivatives are attenuated by $\tau|\bm{V}(\bm{X},t)|$.
The above analysis generalizes to any pixels, any time, any edge orientations.

\subsection{Proof of Optimal Classifier}

The mean of $E_{opt}(\bm{X})$ is $M(\bm{X})$ due to the Bernoulli probability  in \eqref{eq:EPM}. By law of large numbers, \eqref{eq:cost3} converges to \eqref{eq:cost2}. We use \eqref{eq:cost3} in our work because implementation is far simpler than \eqref{eq:cost1} or \eqref{eq:cost2}.

\subsection{Calibration Optimization}

Although this calibration process does not have to run in real-time (since EPM is only used during benchmarking or training), one can speed up the process by appealing to the fact that $\epsilon$ is only used in the final step of Theorem~\ref{thm:Mx} in \eqref{eq:Mx_probability}. Contrast this to $O$, which is used to find $J_t(\bm{X},t)$ in \eqref{eq:J_t}. Hence searching for the optimal $\widehat{O}$ value is more computationally intensive than searching for $\widehat{\varepsilon}$, in general. Hence a cascading of two 1D search algorithm of the form:
\begin{align}\label{eq:mle2}
    (\widehat{\varepsilon},\widehat{O})
    =\arg\max_{O}\left(\max_{\varepsilon}\log P[E|\varepsilon,O]\right),
\end{align}
is more efficient than a literal implementation of a single 2D search algorithm in \eqref{eq:mle2}. 

\subsection{Additional Qualitative Evaluation}

Figure~\ref{tab:additionalImageCompareTable} illustrates how previous denoising algorithms performed on the DVS Optical Flow and IROS18 datasets. IE+TE removes most of the noise, but also removes a large amount of real events. BAF and NN2 allow obvious noise through, but EDnCNN removes a large amount of the noise while retaining the signal in the DVS events.

\end{document}